%% file: main.tex
\documentclass[review]{fcs}

\usepackage{makecell}
\usepackage{balance}
\usepackage{ragged2e}
\usepackage{colortbl}
\usepackage{tcolorbox}
\usepackage{adjustbox}
\usepackage{tikz}
\usepackage[edges]{forest}
\usepackage[switch]{lineno}  %
\usepackage{amsmath,amssymb,amsfonts}
\usepackage{amsthm}
\usepackage{algorithm}
\usepackage{algorithmic}
\usepackage{graphicx}
\usepackage{textcomp}
\usepackage{bm}
\usepackage{subfigure}
\usepackage[american]{babel}
\usepackage{setspace}
\usepackage{enumerate}
\usepackage{multirow}
\usepackage{url}
\usepackage{color}
\usepackage{booktabs}
\usepackage{arydshln}
\usepackage{caption}
\usepackage{CJKutf8}

\title{A Survey of Controllable Learning: Methods and Applications in Information Retrieval}
\author[1]{Chenglei Shen}
\author[1,+]{Xiao Zhang}
\author[1]{Teng Shi}
\author[1]{Changshuo Zhang}
\author[1]{Guofu Xie}
\author[1]{Jun Xu}
\author[2]{Ming He}
\author[2]{Jianping Fan}

\address[1]{Gaoling School of Artificial Intelligence, Renmin University of China, Beijing 100872, China} 

\address[2]{AI Lab at Lenovo Research, Beijing 100085, China}
\corremail{\{chengleishen9,zhangx89\}@ruc.edu.cn}
\fcssetup{
  received = {month dd, yyyy},
  accepted = {month dd, yyyy},
}

\begin{abstract}
Controllability has become a crucial aspect of trustworthy machine learning, enabling learners to meet predefined targets and adapt dynamically at test time without requiring retraining as the targets shift.
We provide a formal definition of controllable learning (CL), and discuss its applications in information retrieval (IR) where information needs are often complex and dynamic.
The survey categorizes CL according to what is controllable (e.g.,  multiple objectives,  user portrait,  scenario adaptation), who controls (users or platforms), how control is implemented (e.g., rule-based method, Pareto optimization, hypernetwork and others), and where to implement control (e.g., pre-processing, in-processing, post-processing methods). 
Then, we identify challenges faced by CL across training, evaluation, task setting, and deployment in online environments. Additionally, we outline promising directions for CL in theoretical analysis, efficient computation, empowering large language models, application scenarios and evaluation frameworks.
\end{abstract}

\begin{document}

\input{sections/1_introduction}

\input{sections/2_formulation}

\input{sections/3_control_ir}

\input{sections/4_methods}

\input{sections/5_resources}

\input{sections/6_challenge}

\input{sections/7_conclusion}

\section*{Acknowledgement}
This work was partially supported by the National Natural Science Foundation of China (No. 62376275, 62472426). Work partially done at Beijing Key Laboratory of Research on Large Models and Intelligent Governance, and Engineering Research Center of Next-Generation Intelligent Search and Recommendation, MOE. Supported by fund for building world-class universities (disciplines) of Renmin University of China.





\bibliographystyle{fcs}
\bibliography{ref}

\begin{biography}{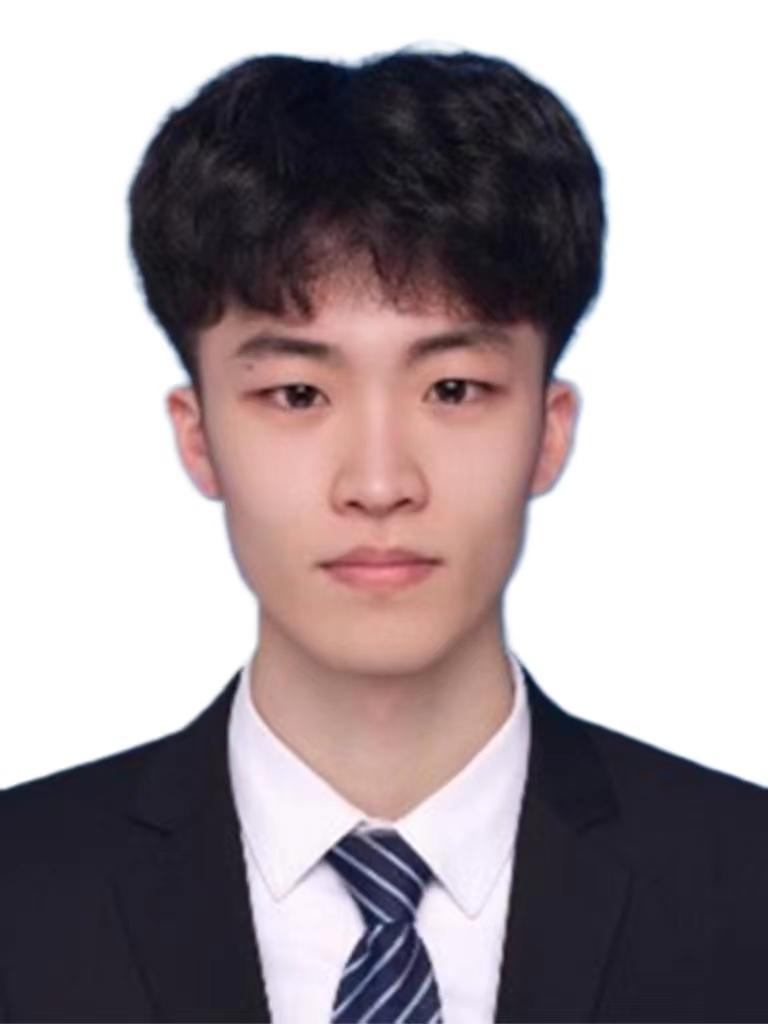}
Chenglei Shen is currently pursuing his Ph.D. degree
of  Artificial Intelligence at Gaoling School of Artificial intelligence, Renmin University of China. His current research interests mainly include controllable learning, information retrieval, and large language models.
\end{biography}

\begin{biography}{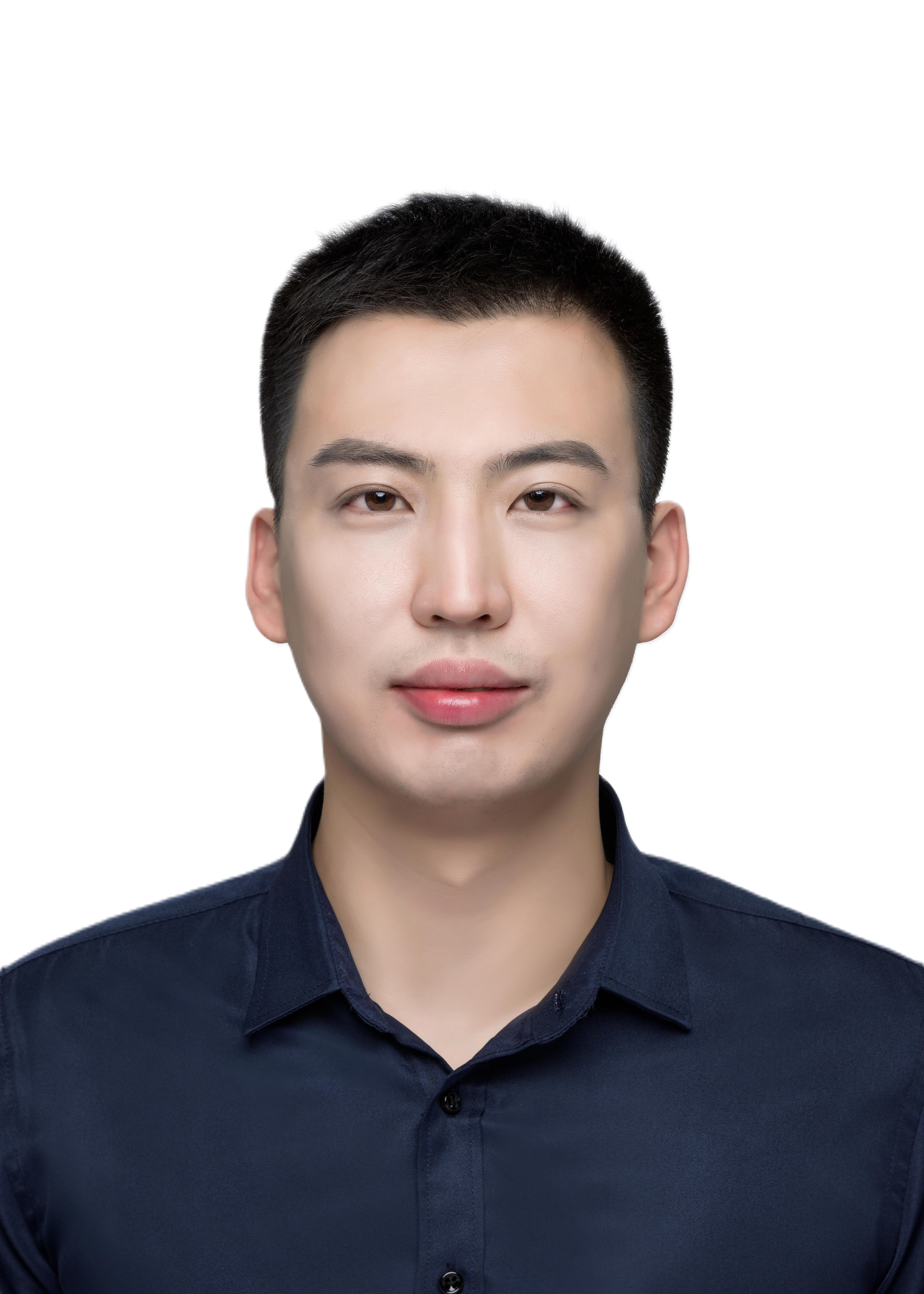}
Xiao Zhang is an associate professor at Gaoling School of Artificial Intelligence, Renmin University of China. His research interests include online learning, trustworthy machine learning, and information retrieval. He has published over 60 papers on top-tier conferences and journals in artificial intelligence, e.g., NeurIPS, ICML, KDD, SIGIR, AAAI, IJCAI, ICLR, WWW, VLDB, etc.
\end{biography}

\begin{biography}{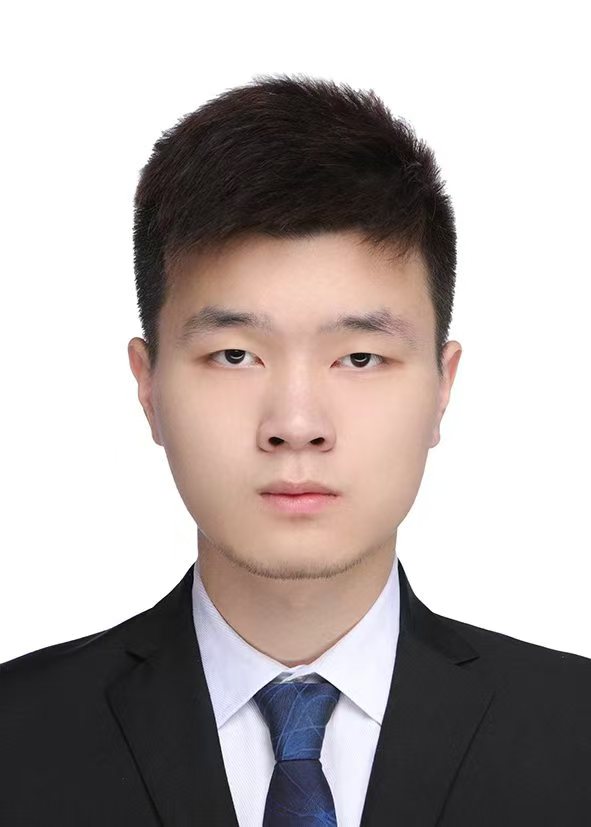}
Teng Shi
is currently pursuing his Ph.D. degree
of  Artificial Intelligence at Gaoling School of Artificial intelligence, Renmin University of China. His current research interests mainly include  information retrieval and recommender systems.
\end{biography}

\begin{biography}{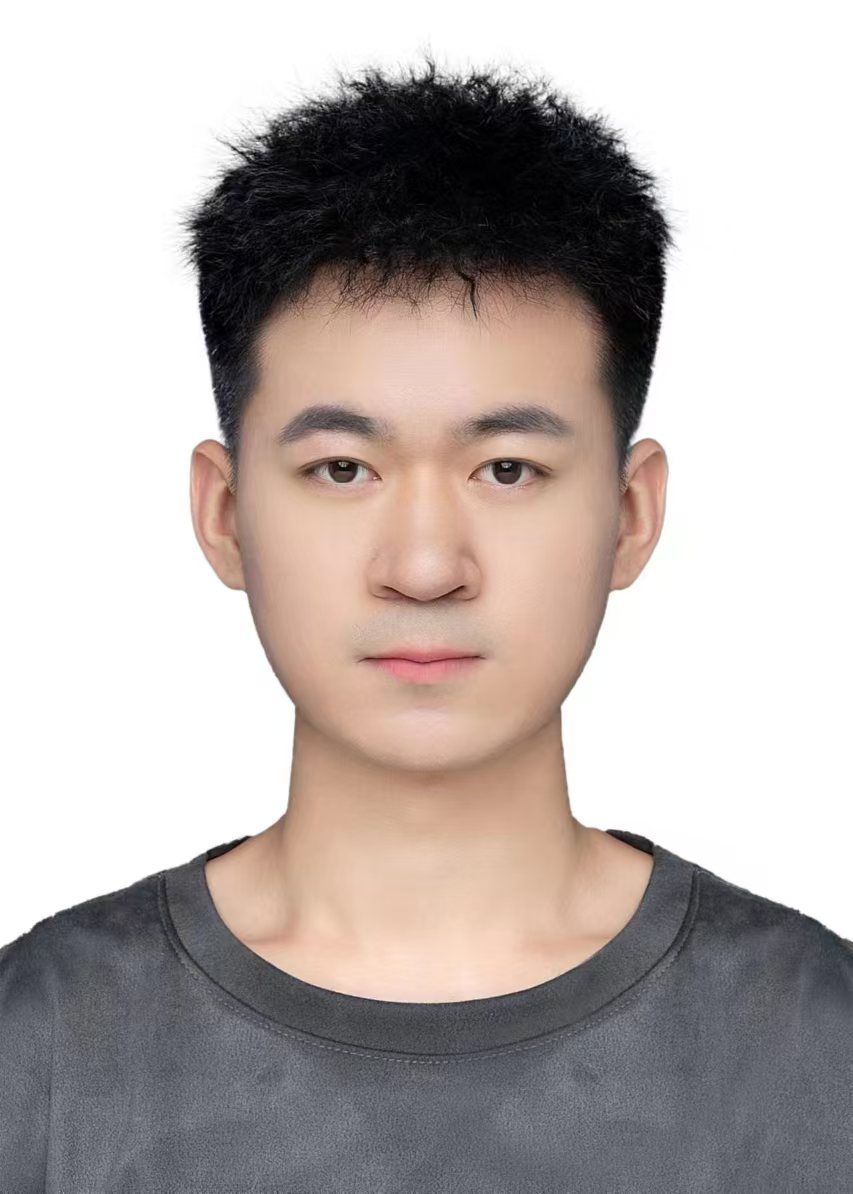}
Changshuo Zhang
is currently pursuing his Master's degree in Artificial Intelligence at Gaoling School of Artificial Intelligence, Renmin University of China. His current research interests mainly include information retrieval and recommender systems.
\end{biography}

\begin{biography}{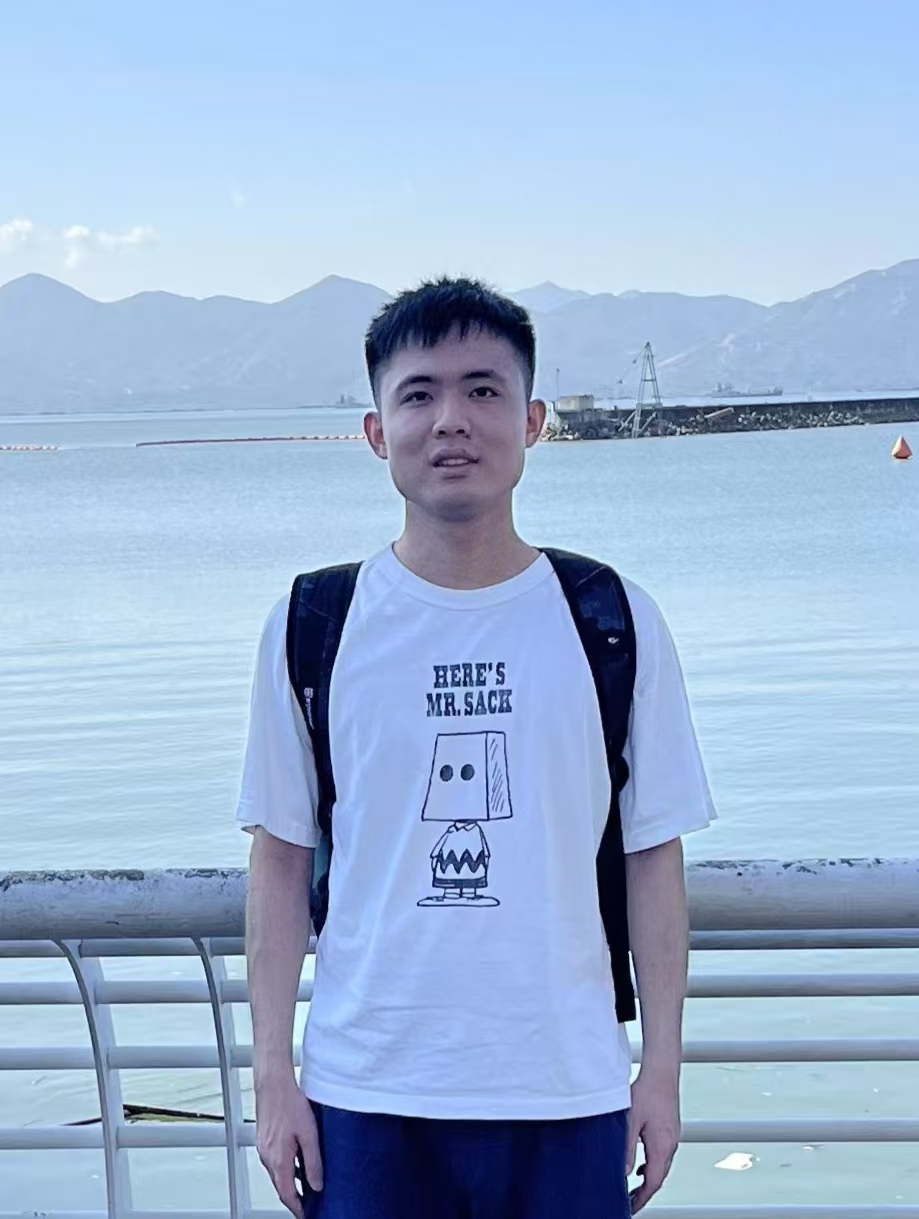}
Guofu Xie
is currently pursuing his Ph.D. degree
of  Artificial Intelligence at Gaoling School of Artificial intelligence, Renmin University of China. His current research interests mainly include information retrieval and large language models.
\end{biography}

\begin{biography}{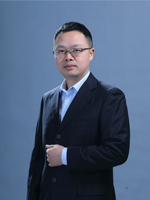} Jun Xu is a Professor at Gaoling School of Artificial Intelligence, Renmin University of China. His research interests focus on learning to rank. He has published more than 100 papers in international conferences (e.g., SIGIR, WWW) and journals (e.g., TOIS, JMLR). He serves as SPC for SIGIR, WWW, AAAI, and ACML, Editor of JASIST. He has won the Best Paper Honorable Mention in SIGIR (2024), WWW 2023 spotlight/Best Paper Nominee, and Test of Time Award Honorable Mention in SIGIR (2019).
\end{biography}

\begin{biography}{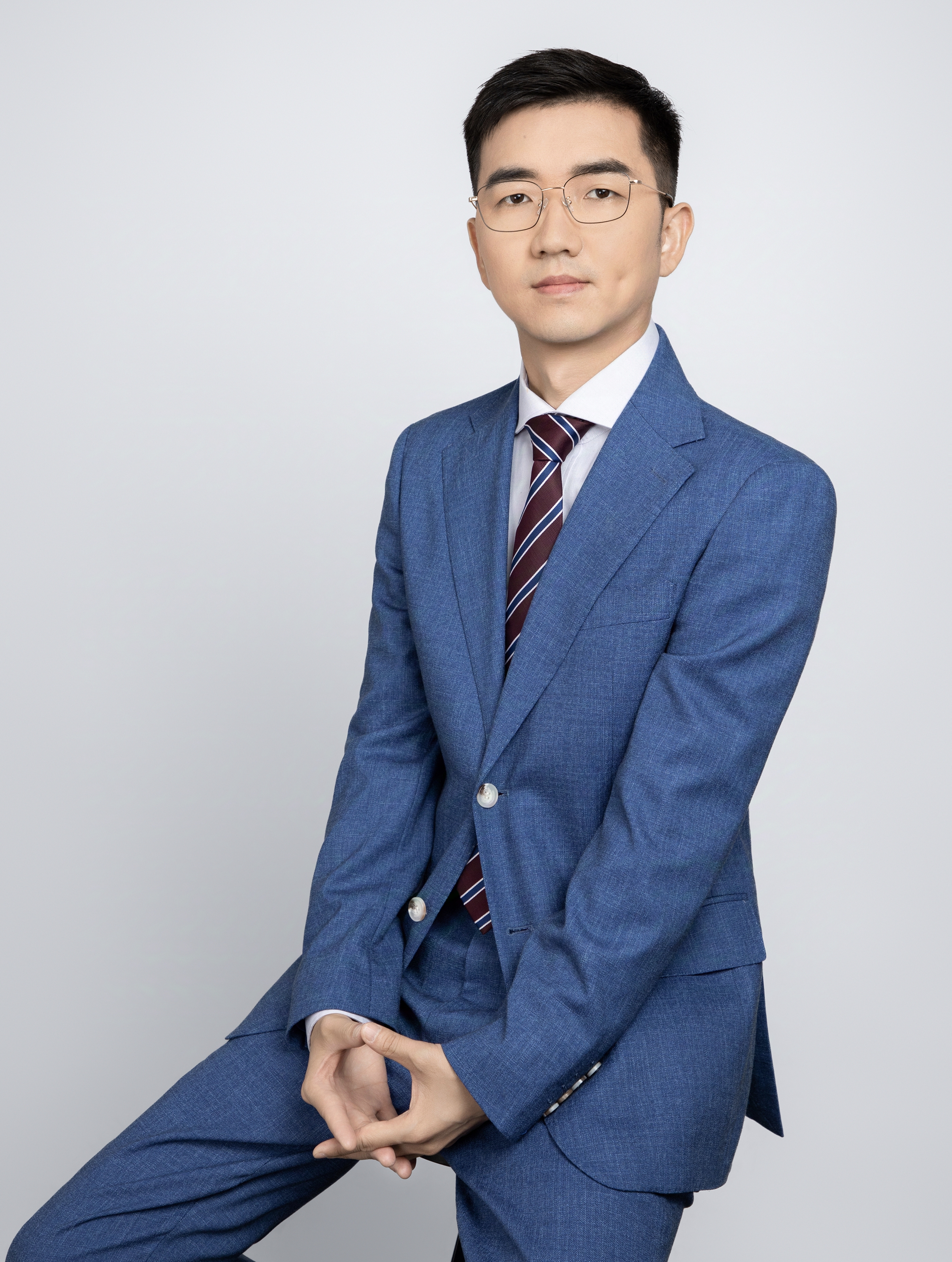} Ming He received his Ph.D. degree in computer science from University of Science and Technology of China, Hefei, in 2018; received the title of senior engineer of Chinese Academy of Sciences, Beijing, in 2022. He is currently an advisory researcher with the AI Laboratory, Lenovo Research, Beijing. His current research interests include recommendation systems, AI Agent and decision intelligence. He has published 20+ papers in refereed journals and conference proceedings, e.g., TIP, KDD and OMEGA. He received the KSEM 2018 Best Research Paper Award and submitted 120+ patents to Chinese National Patent Office.
\end{biography}

\begin{biography}
{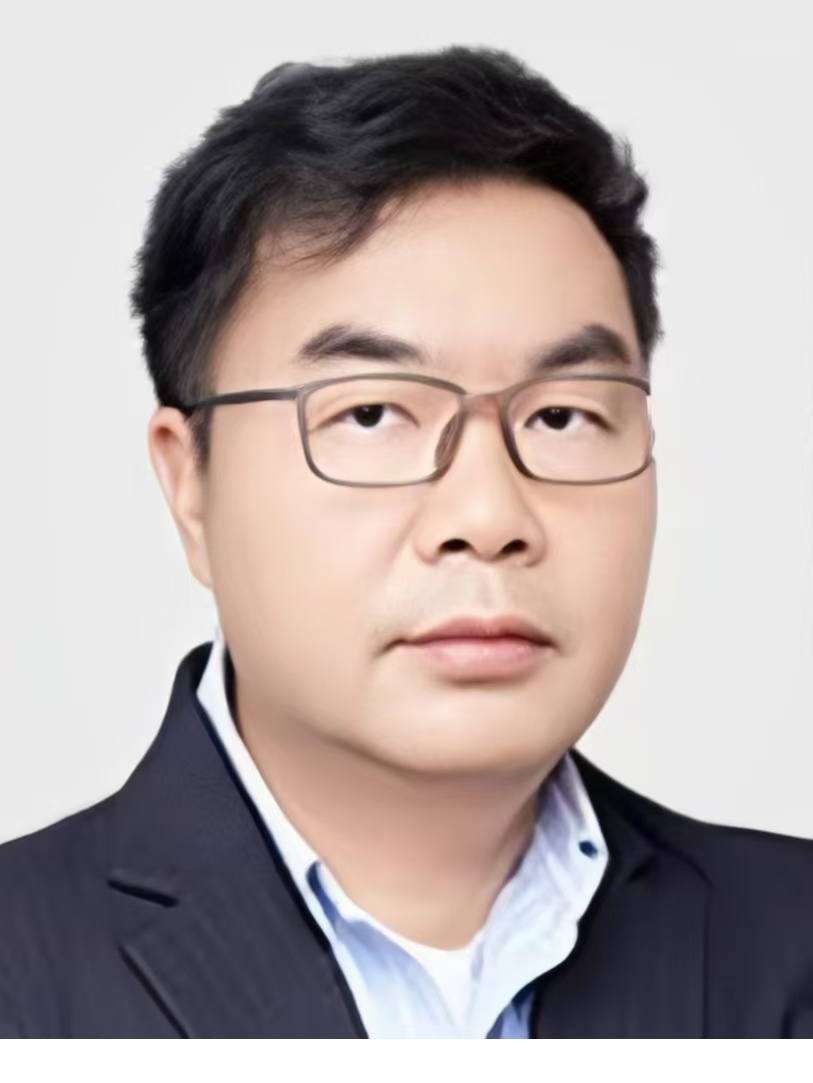} 
Jianping Fan received his Master's degree in theoretical physics from Northwest University (Xi'an, China) in 1994, and his PhD degree in optical storage and computer science from Shanghai Institute of Optics and Fine Mechanics of Chinese Academy of Sciences in 1997. From 1997 to 1998, Dr. Fan was a Researcher at Fudan University (Shanghai, China). From 1998 to 1999, he was a Researcher with Japan Society for the Promotion of Science and Osaka University (Japan). From 1999 to 2001, he was a Postdoctoral Researcher in the Department of Computer Science of Purdue University (West Lafayette, USA). His research interests include image/video privacy protection, computer vision, statistical machine learning and deep learning. 
\end{biography}

\end{document}

%% file: sections/1_introduction.tex
\section{Introduction}
\input{fig/FCS_overview_tree_fig1}



Trustworthiness in machine learning is becoming increasingly important, with considerations such as controllability, fairness, privacy, and interpretability gaining prominence \cite{toreini2020relationship,eshete2021making,wang2023trustworthy,zhang2024model}.
In November 2023, the Bletchley Declaration , adopted at the first Global Artificial Intelligence Safety 
Summit, emphasized that:``\emph{AI should be designed, developed, deployed, and used, in a manner that is safe, in such a way as to be human-centric, trustworthy and responsible\ldots Substantial risks may arise from potential intentional misuse or unintended issues of control relating to alignment with human intent.}'' 
In October 2023, the Cyberspace Administration of China released the Global AI Governance Initiative, which emphasized the need to: ``\emph{\ldots make AI technologies more secure, reliable, controllable, and equitable, \ldots ensure that AI always remains under human control}.''  
In this paper, we explore the topic of controllable AI, with a particular focus on the controllability of machine learning methods and their practical applications. 
Norbert Wiener, the originator of cybernetics, foresaw the  ethical challenge of learning 
 machine over sixty years ago when he stated~\cite{wiener1960some}
``\emph{As machines learn they may develop unforeseen strategies at rates that baffle their programmers.}'' 
Additionally, Wiener explained the significance of controllability in learning machines~\cite{wiener1960some}: ``\emph{If we use, to achieve our purposes, a mechanical agency with whose operation we cannot efficiently interfere once we have started it, because the action is so fast and irrevocable that we have not the data to intervene before the actions complete, then we had better be quite sure that the purpose put into the machine is the purpose which we really desire and not merely a colorful imitation of it, etc.}'', emphasizing two fundamental aspects of AI controllability: (1)~Controllable learning (CL) machine needs to ensure that its predictions/decisions meet the target of the AI user; (2)~Intervening in a learning machine after its deployment is a crucial but challenging task.


Under this insight and context, we present an analysis of the feasibility and necessity of CL from three perspectives: the technical foundations of controllability, the practical value of CL in information retrieval, and the landscape of commercial deployment.

\subsection{Technical Foundations of Controllable Learning (CL)}

Controllability has been extensively discussed in the field of \emph{generative} machine learning models, also known as controllable generation. For instance, in text generation~\cite{keskar2019ctrl,Hu2017Toward,Zhou2023Controlled} or visual generation~\cite{epstein2023diffusion,li2024blip,chen2023control}, the AI models take a given task description  (often referred to as a ``prompt'') as input and generates content that aligns with that description.
However, in the realm of \emph{discriminative} machine learning models and applications, there lacks a unified definition and in-depth discussion of controllability. The current popular prompt-based controllable generation methods can be seen as a form of pre-processing, where, under fixed model parameters, changing input features allows the model to meet the target of the AI user.
Existing theoretical results suggest that controlling model parameters may lead to better convergence than controlling model inputs when given task requirements \cite{galanti2020modularity}, revealing that achieving controllability in machine learning involves various methods beyond prompt-based approaches. 
Therefore, controllable learning (CL) is a crucial category worthy of comprehensive review and exploration in trustworthy machine learning, ensuring that learning models meet user expectations and adhere to ethical standards. 

\subsection{Practical Value of CL in Information Retrieval}

Building on the technical foundations of CL, it is important to examine how these principles translate into practical benefits within real-world machine learning applications. One such domain is Information Retrieval (IR). Specifically, in the context of learning-based Information Retrieval (IR) applications, users actively search for information or passively receive recommendations, both aimed at fulfilling their information-seeking needs \cite{garcia2011information,xu2018deep,shi2024unisar,shi2025unified,zhang2024qagcf}.
In these applications, CL enables models to adapt to to diverse task description (e.g, users' information needs) at test time without extensive retraining, delivering personalized and relevant results. 
Given the above landscape, this survey attempts to provide a formal definition of CL, classify and discuss its application paradigms in information retrieval. 
In summary, this survey focuses on the topic of CL, which aims to find a learner capable of adapting to different task requirements without the need for retraining, thereby meeting the desired task targets of the AI users.

\subsection{Landscape of Commercial Deployment}

From a commercial perspective, 
the ongoing evolution of large-scale machine learning models, including advanced large language models~\cite{zhao2023survey} like ChatGPT, has prompted a transition in the deployment of machine learning (ML) algorithms, moving away from the conventional Software-as-a-Service (SaaS)~\cite{cusumano2010cloud} towards Model-as-a-Service (MaaS)~\cite{gan2023model}. 
MaaS provides users with access to trained ML models via user-friendly interfaces, such as application programming interfaces (APIs). More specifically, MaaS allows users to access the trained models by calling APIs without the need to undertake the significant costs associated with model training and updating.  
MaaS empowers both companies and individuals to leverage the powerful capabilities of large models. However, it also poses new challenges for the controllability of ML models. Specifically, it raises the question of how to make ML models recognize the requirements of different downstream tasks and output personalized results that align with task targets without the necessity to retrain the model. 

In the context of MaaS, the demand for controllability of ML models in information retrieval applications becomes more pronounced. This is attributed to the inherent complexity of information needs in IR tasks, where precise and tailored results are critical.   
In traditional settings, such as keyword-based search engines, the retrieval process is relatively straightforward, relying on predefined rules and algorithms. However, with the advent of MaaS and the utilization of large language models, the landscape has evolved. Now, users expect finer-grained and context-aware responses based on their information needs~\cite{maps,xie2025bone}. 
This heightened expectation emphasizes the importance of controllability of ML models. Specifically, there is a need for mechanisms that enable individuals or platforms to easily express and convey their specific requirements for IR tasks. 
For example, in an e-commerce setting, a user searching for ``black T-shirt'' may expect different results based on varying weights of evaluation metrics 
(such as favoring diversity to simultaneously expose clothing that pairs well with black T-shirts), or setting specific filters to exclude irrelevant content (such as not exposing previously purchased T-shirts). Achieving this level of controllability necessitates sophisticated ML models capable of dynamically adjusting model parameters or outputs based on task requirements, without extensive retraining. 
Therefore, in the realm of IR, MaaS not only provides unprecedented access to powerful ML models but also underscores the significance of controllability and customization in addressing the diverse needs of users across different domains and applications.


\subsection{Comparison with Existing Surveys} 
This section aim to state the distinctions between this survey and existing surveys. To the best of our knowledge, there is a notable lack of comprehensive surveys specifically dedicated to controllable learning in information retrieval. To date, only one survey from 2017 has touched upon the concept of ``user control'' as a keyword~\cite{jannach2017user}. Therefore, we compare our work with surveys in closely related areas, including Trustworthy Information Retrieval~\cite{ge2022survey}, and Explainable Information Retrieval~\cite{zhang2020explainable, anand2022explainable}

To begin with, the survey on \textbf{user control}~\cite{jannach2017user} focuses on the instantaneous nature of user interests. It highlights the need for mechanisms that allow users to take control of recommendations when the system's assumptions about their preferences are inaccurate or outdated. However, the survey exhibits significant limitations. It considers controllability exclusively from the user's perspective, neglecting the role of platform-side control. Moreover, it merely summarizes existing work in terms of interaction forms, lacking rigorous definitions and in-depth technical analysis.
Similarly, the survey on \textbf{trustworthy} information retrieval~\cite{ge2022survey} primarily focuses on the security aspects of privacy preservation in recommender systems. Nevertheless, it also addresses controllability, categorizing it into two distinct types: explicit controllability and implicit controllability. The former allows users explicitly edit or update the user preferences. The latter means that users could indirectly fine-tune their preferences when dynamically interacting with the recommender systems through re-ranking, modifying historical information, and so on. However, this survey lacks a systematic summary of the definition of controllability in machine learning and the methods by which controllability can be achieved. Moreover, there may be other possibilities for controllability in a broader sense, such as platform-side controllability or controllability with multiple objectives, which are not thoroughly defined and categorized. Additionally, some surveys on \textbf{explainable} information retrieval (IR) have expressed concerns related to controllability. However, there remains a significant distinction between explainability and controllability. First, explainability primarily focuses on making the  embedding-based retrieval or recommendation models transparent and interpretable. In contrast, the notion of controllability discussed in this work emphasizes enabling retrieval or recommendation models to generate results under control, in accordance with given specific requirements. For instance, a recent survey on explainable IR~\cite{zhang2020explainable, anand2022explainable} paraphrases explainability as the ability to provide users or system designers with reasons clarifying why certain items are retrieved or recommended. Second, explainability can be regarded as a preliminary step toward achieving controllability. By offering such explanations, explainable IR facilitates human understanding of algorithmic decisions—whether the human is an end user or a system designer. This, in turn, enhances key aspects such as transparency, persuasiveness, effectiveness, and controllability. In summary, the key scientific challenge in the field of explainable IR is how to make the process and results of IR systems interpretable. Moreover, strong explainability can promote the research and development of CL.

Additionally, these surveys are also relatively early, and many new controllable techniques (such as hypernetworks~\cite{chauhan2023brief}, large language models~\cite{zhao2023survey}, etc.) have emerged since then.  Considering the limitations of existing surveys, in this survey, we provide a formal definition of CL, and discuss CL and its applications in information retrieval.

\subsection{Contribution}
The current landscape of research and application deployment in the aforementioned areas motivates us to write a comprehensive survey on CL. 
As shown in Figure~\ref{fig:overview_tree}, this survey will encompass various aspects including the formal definition of CL, methods and applications in information retrieval, evaluations, challenges, and future directions. 
\begin{itemize}
    \item \emph{Formal definition of CL.} The survey will begin by providing a clear and concise formal definition of CL. 
    \item \emph{Applications of CL in IR.} We will then delve into the applications of CL in the field of IR, 
    exploring who controls, what is controllable, and how control is implemented in IR systems. 
    \item  \emph{Future directions and challenges.} Finally, the survey will discuss the future directions and potential challenges in the field of CL. 
\end{itemize}

Overall, the survey on CL will serve as a valuable resource for students, researchers, practitioners, and policymakers interested in understanding the principles, applications, and future prospects of this important area of trustworthy machine learning.

%% file: fig/FCS_overview_tree_fig1.tex
\definecolor{fill_0}{RGB}{160, 190, 230}  
\definecolor{fill_1}{RGB}{240, 195, 175}  
\definecolor{fill_2}{RGB}{240, 245, 235}
\definecolor{fill_3}{RGB}{250, 246, 235}

\definecolor{draw_0}{RGB}{60, 90, 120}    
\definecolor{draw_1}{RGB}{120, 60, 60}    
\definecolor{draw_2}{RGB}{90, 120, 60}    
\definecolor{draw_3}{RGB}{120, 110, 60}   

\definecolor{fill_leaf}{RGB}{248, 248, 248}
\definecolor{draw-leaf}{RGB}{135, 135, 135}

\tikzstyle{my-box}=[
    rectangle,  
    draw=draw-leaf,
    rounded corners,
    text opacity=1,
    minimum height=2.5em,
    minimum width=5em,
    inner sep=2pt,
    align=center,
    fill opacity=.5,
    line width=0.95pt,
]
\tikzset{
    leaf/.style={
        my-box,
        minimum height=4.5em,
        fill=fill_leaf,
        text=black,
        align=left,
        font=\footnotesize,
        inner xsep=2pt,
        inner ysep=4pt,
        line width=1pt
    }
}

\begin{figure*}[!t]
\centering
\begin{adjustbox}{width=1.0\textwidth}
\begin{forest}
forked edges,
for tree={
    grow=east,
    reversed=true,
    anchor=base west,
    parent anchor=base east, 
    child anchor=base west, 
    base=center,
    font=\small,
    rectangle,
    draw=draw-leaf,
    rounded corners,
    align=left,
    text centered,
    minimum width=5em,
    edge+={darkgray, line width=1pt},
    s sep=3pt,
    inner xsep=2pt,
    inner ysep=3pt,
    line width=1.2pt,
    ver/.style={rotate=90, child anchor=north, parent anchor=south, anchor=center},
},
where level=0{text width=13em,fill=fill_0,draw=draw_0,font=\normalsize,}{},
where level=1{text width=9.5em,fill=fill_1,draw=draw_1,font=\normalsize,}{},
where level=2{text width=10.4em,fill=fill_2,draw=draw_2,font=\normalsize,}{},
where level=3{text width=10.4em,fill=fill_3,draw=draw_3,font=\normalsize,}{},
where level=4{font=\LARGE,}{},
[
    Controllable Learning \\(CL) , minimum height=3em
    [ 
         Formulation of CL\\ (Sec.~\ref{sec:formulation_of_CL})
        [
            Definition of CL \\(Sec.~\ref{sec:formulation_of_CL})
            [
                \parbox{37em}{Define a task requirement triplet, $\mathcal{T} = \{ \bm s_{\mathrm{desc}}, \bm s_{\mathrm{ctx}}, \bm s_{\mathrm{tgt}}\}$, where $\bm s_{\mathrm{desc}} \in \mathcal{D}_{\mathrm{desc}}$ represents the task description, $\bm s_{\mathrm{ctx}} \in \mathcal{D}_{\mathrm{ctx}}$ represents the context related to the task, and $\bm s_{\mathrm{tgt}} \in \mathcal{D}_{\mathrm{tgt}}$ represents the task target.}
                ,leaf, text width=31em
            ]
        ]
    ]
    [
         Taxonomy of CL \\ (Sec.~\ref{sec:taxonomy})
         [
            What is controllable\\ (Sec.~\ref{sec:what_to_control})
            [
                Multi-Objective Control \\ (Sec.~\ref{sec:multi-faced_retrieval_objectives})
                [
                    \parbox{36em}{Taking several objectives (metrics) as control targets, these work includes  ComiRec~\cite{cen2020controllable}, UCRS~\cite{wang2022user}, CMR~\cite{chen2023controllable}, PadiRec~\cite{shen2024generating}, RecLM-gen~\cite{lu2024aligning}}, leaf, text width=31em
                ]
            ]
            [
                User Portrait Control\\(Sec.~\ref{sec:historical_behavior_control})
                [
                    \parbox{36em}{Taking historical behavior as control targets , these work includes LACE~\cite{mysore2023editable}, LFRQE~\cite{wang2024would}, Grapevine~\cite{rahdari2021connecting}, TEARS~\cite{penaloza2024tears}, CMBR~\cite{gou2024controllable}, UCR~\cite{tan2023user}}, leaf, text width=31em
                ]
            ]   
            [
                Scenario Adaptation \\Control (Sec.~\ref{sec:controllable_environmental_adaptation})
                [
                    \parbox{36em}{Taking environment adaptation as target, these work includes HyperBndit\cite{shen2023hyperbandit}, Hamur~\cite{li2023hamur}, HyperBandit~\cite{shen2023hyperbandit}, PEPNet~\cite{chang2023pepnet}}, leaf, text width=31em
                ]
            ]   
        ]
        [
             Who controls \\ (Sec.~\ref{sec:who_controls})
             [
                User-centric control \\ (Sec.~\ref{sec:user_centric_control})
                [
                    \parbox{36em}{Prior profile modification which empowers users to actively shape their recommendation experience: UCRS~\cite{wang2022user}, IFRQE~\cite{wang2024would}, LACE~\cite{mysore2023editable}, Supervised $\beta$-VAE~\cite{nema2021disentangling}, LP~\cite{luo2020latent}, TEARS~\cite{penaloza2024tears}, LangPTune~\cite{gao2024end},UCRS~\cite{wang2022user}, GOMMIR~\cite{wu2023goal}, Promptriever~\cite{weller2024promptriever}, InstructAgent~\cite{xu2025instructagent}}
                    , leaf, text width=31em
                ]
             ]
             [
                Platform-mediated control \\ (Sec.~\ref{sec:platform_mediated_control})
                [
                    \parbox{36em}{Impose algorithmic adjustments and policy-based constraints on the recommendation process: ComiRec~\cite{cen2020controllable}, CMR~\cite{chen2023controllable}, CCDF~\cite{zhang2024practical} and ~\cite{li2023breaking}, Padirec~\cite{shen2024generating}, SAMD~\cite{huan2023samd}, HyperBandit~\cite{shen2023hyperbandit}, DTRN~\cite{liu2023deep}
                    }
                    , leaf, text width=31em
                ]
             ] 
        ]    
    ]
    [
        How control is \\implemented  (Sec.~\ref{sec:implement_control})
        [
             CL Techniques \\ (Sec.~\ref{sec:CL_technology})
            [
                Rule-Based Control\\ (Sec.~\ref{sec:rule_based_control})
                [
                \parbox{36em}{Rule-based post-processing techniques: \cite{parikh2023information}, \cite{khosrobeigi2020rule}, \cite{kang2013using}, \cite{nandy2022achieving}, \cite{le2023combining} and \cite{antikacioglu2017post}.}
                , leaf, text width=31em
                ]
            ]
            [
                Pareto optimization\\ (Sec.~\ref{sec:pareto_optimization})
                [
                \parbox{36em}{Achieving controllability by using Pareto Optimization: PHN-HVI~\cite{hoang2023improving}, PHNs~\cite{navon2020learning}, ~\cite{ruchte2021scalable}, ~\cite{lin2020controllable}, ~\cite{ma2020efficient}, ~\cite{lin2019pareto}, ~\cite{ribeiro2014multiobjective} and ~\cite{ribeiro2012pareto}.}
                , leaf, text width=31em
                ]
            ]
            [
                Hypernetwork\\ (Sec.~\ref{sec:parameter_control})
                [
                    \parbox{36em}{Hypernetworks based methods: Hamur~\cite{li2023hamur}, CMR~\cite{chen2023controllable}, Hyperprompt~\cite{he2022hyperprompt}, HyperBandit~\cite{shen2023hyperbandit}, PadiRec~\cite{shen2024generating}, Hypencoder~\cite{killingback2025hypencoder}, \cite{galanti2020modularity}, \cite{hoang2023improving}, and \cite{navon2020learning}}
                , leaf, text width=31em
                ]
            ]
        ]
        [
            Where to control\\ (Sec.~\ref{sec:where_to_control})
            [
                Pre-processing\\  methods(Sec.~\ref{sec:pre_processing_methods})
                [
                \parbox{36em}{Including data manipulation methods and feature engineering methods: UCRS~\cite{wang2022user},  LACE~\cite{mysore2023editable}, ~\cite{wang2024would} and MetaLens~\cite{schafer2002meta}. TART~\cite{asai2022task}, Instructor~\cite{su2022one}, FollowIR~\cite{weller2024followir}, InstructIR~\cite{oh2024instructir}}
                , leaf, text width=31em
                ]
            ]
            [
                In-processing \\methods(Sec.~\ref{sec:in_processing_methods})
                [
                \parbox{36em}{Including regularization and constrained optimization methods:CMR~\cite{chen2023controllable}, CCDF~\cite{zhang2024practical}, HyperBandit~\cite{shen2023hyperbandit}}
                , leaf, text width=31em
                ]
            ]
            [
                Post-processing \\methods(Sec.~\ref{sec:post_processing_methods})
                [
                 \parbox{36em}{Including reranking and result diversification methods: ComiRec~\cite{cen2020controllable} and MMR~\cite{carbonell1998use}}
                , leaf, text width=31em
                ]
            ]
        ]
    ]
    [
        Evaluations for CL \\ in IR (Sec.~\ref{sec:evaluation})
        [
            Metrics (Sec.~\ref{sec:metric}), minimum height=3em
            [
                Single-objective\\ metrics(Sec.~\ref{sec:Single-objective metrics})
                [
                 \parbox{36em}{NDCG~\cite{jarvelin2002cumulated}, Recall, ~Precision, ~Hit Rate, ~$\alpha$-NDCG~\cite{clarke2008novelty}, ERR-IA~\cite{yan2021diversification}, Coverage, Iso-Index.}
                , leaf, text width=31em
                ]
            ]
            [
                Multi-objective \\metrics(Sec.~\ref{sec:Multi-objective optimization metric})
                [
                 \parbox{36em}{Typical Multi-objective metrics include 4 main groups~\cite{audet2021performance}. However, only the Hypervolume indicator~\cite{zitzler1999evolutionary} is most suitable in the current discussion context.}
                 , leaf, text width=31em
                ]
            ]
        ]
        [
            Datasets (Sec.~\ref{sec:dataset}), minimum height=3em%
            [
                \parbox{36em}{Amazon~\cite{ni2019justifying,hou2024bridging}, Ali\_Display\_Ad\_Click~\cite{zhou2018deep}, UserBehavior~\cite{cen2020controllable}, MovieLens, MS~MARCO~\cite{nguyen2016ms}}, leaf, text width=31em
            ]
        ]
    ]
    [
        Challenges in CL\\ for IR (Sec.~\ref{sec:challenges})
        [
            Balancing difficulty \\(Sec.~\ref{sec:balancing_difficulty})
            [
                 \parbox{36em}{Pursuing controllability often leads to a trade-off, potentially compromising performance or other user-centric optimization metrics and adversely impacting accuracy or user experience.},leaf, text width=31em
            ]
        ]
        [
            Absence of evaluation \\(Sec.~\ref{sec:absence_of_evaluation})
            [
                 \parbox{36em}{The assortment of perspectives on controllability and the consequent need for tailored evaluation metrics prevents direct methodological comparisons and hinder the progression of the field.}, leaf, text width=31em
            ] 
        ]
        [
            Descriptions setting\\ challenges (Sec.~\ref{sec:challenge:3})
            [
                 \parbox{36em}{The task description serves as the instructions given by humans to the learner. A crucial issue is how to set the task target and transform it into a human understandable and precise description.},leaf, text width=31em
            ] 
        ]
        [
            Online challenges  \\(Sec.~\ref{sec:online_challenges})
            [
                \parbox{36em}{Scalability in realworld IR systems, particularly those dealing with streaming data and requiring continuous learning, is a formidable challenge.}, leaf, text width=31em
            ] 
        ]
    ]
    [
        Future directions of \\CL (Sec.~\ref{sec:future_direction})
        [
            Several categories\\(Sec.~\ref{sec:future_direction}) \\
            [
                 \parbox{36em}{Including theoretical analyses of CL, controllable decision-making models, empowering LLM-based AIGC through CL. cost-effective control learning mechanisms, CL for multi-task switching, demand for resource and metrics.},leaf, text width=31em
            ]
        ]
    ]
]
\end{forest}
\end{adjustbox}

\vspace{1mm}
\caption{Overview of the survey of CL. This survey includes formulation of CL, taxonomy of CL, implementation methods of CL, how to evaluate the effectiveness of CL in information retrieval (IR) applications, and challenges of applying CL in IR.}
\label{fig:overview_tree}
\end{figure*}

%% file: sections/2_formulation.tex
\section{Foundations of Controllable Learning (CL)}
\label{sec:formulation_of_CL} 
In this section, we first give a formal definition of CL in terms of task requirements, and describe the procedure of CL.



The concept of CL is manifested across various AI research directions, yet there is currently lack of a unified formal definition. To explicitly express the ingredients of CL and unify existing CL methods, we first define CL in terms of task specifications. 
In brief, CL is the ability to find a learner that can adapt to different task requirements without the need for retraining,  thereby meeting the desired task targets, as detailed in Definition~\ref{def:CL:first}.

\begin{definition}[Controllable Learning (CL)]
\label{def:CL:first}
Define a task requirement triplet $\mathcal{T} = \{ \bm s_{\mathrm{desc}}, \bm s_{\mathrm{ctx}}, \bm s_{\mathrm{tgt}}\} \in \Gamma$, where $\bm s_{\mathrm{desc}} \in \mathcal{D}_{\mathrm{desc}}$ represents the task description, $\bm s_{\mathrm{ctx}} \in \mathcal{D}_{\mathrm{ctx}}$ represents the context related to the task, and $\bm s_{\mathrm{tgt}} \in \mathcal{D}_{\mathrm{tgt}}$ represents the task target.
Given an input space $\mathcal{X}$ and an output space $\mathcal{Y}$, 
for a learner $f: \mathcal{X} \rightarrow \mathcal{Y}$, 
controllable learning (CL) aims to find a control function $h$ that maps the learner $f$, the task description $\bm s_{\mathrm{desc}} \in \mathcal{T}$, and the context $\bm s_{\mathrm{ctx}} \in \mathcal{T}$ to a new learner $f_{\mathcal{T}}$ that fulfills the task target $\bm s_{\mathrm{tgt}} \in \mathcal{T}$, i.e, 
$$
     f_{\mathcal{T}} = h(f, \bm s_{\mathrm{desc}}, \bm s_{\mathrm{ctx}}).
$$
The integration of the learner $f$ and the control function $h$ is called a controllable learner. 
Moreover, upon receiving a new task requirement $\mathcal{T}' \in \Gamma$ at test time
, the control function $h$ should be capable of outputting a new learner $f_{\mathcal{T}'}$ without the need for model retraining,  ensuring that $f_{\mathcal{T}'}$ satisfies the task target $\bm s'_{\mathrm{tgt}} \in \mathcal{T}'$.

\end{definition}

Unlike domain adaptation or transfer learning\cite{pan2009survey,weiss2016survey,zhuang2020comprehensive}, CL adjusts the model  adaptively in the face of new task requirements, eliminating the need for retraining for new tasks during the deployment phase. 
More specifically, in the context of a controllable learner during the testing phase, adjustments based on the task description can be categorized as follows: altering model inputs qualifies as a pre-processing method, modifying model parameters constitutes an in-processing method, and changing model outputs can be regarded as a post-processing method.
However, it also presents challenges during the training phase in CL, particularly in guiding the control function to identify and adapt to various task requirements effectively. 

For the task requirement in Definition~\ref{def:CL:first}, the task target $\bm s_{\mathrm{tgt}}$ can be considered as the ideal quantitative metric that the controllable learner aims to achieve. The task description $\bm s_{\mathrm{desc}}$ is the specific representation of task target $\bm s_{\mathrm{tgt}}$ that can be perceived by the control function $h$. Here is an example of task description $\bm{s}_{\mathrm{des}}$:
Here is an example to illustrate the task target $\bm s_{\mathrm{tgt}}$ and the task description $\bm{s}_{\mathrm{des}}$:

\begin{example}
In the e-commerce platform, the platform often have requirements for both accuracy and diversity in the output results, and their preference between the two may shift depending on business objectives. In such scenarios, task target  $\bm{s}_{\mathrm{tgt}}$ can represent an expected performance target with respect to both accuracy and diversity (e.g., placing greater emphasis on diversity).
Following this scenario, $\bm{s}_{\mathrm{desc}}$ can be represented either as a vector containing the weights of accuracy and diversity  (e.g., $[0.4,0.6]$). Thus, at test time, platforms can express their preferences for diversity and accuracy through $\bm{s}_{\mathrm{desc}}$, which serves as the input to the control function $h$, thereby producing a learner $f$ that satisfies the target $\bm{s}_{\mathrm{tgt}}$.
\end{example}
The context $\bm s_{\mathrm{ctx}}$ in task requirements can consist of features like historical data and the current user's profile, offering controllable learners more background knowledge for prediction.

\begin{figure*}[t]
    \centering
    \includegraphics[width=0.75\textwidth]{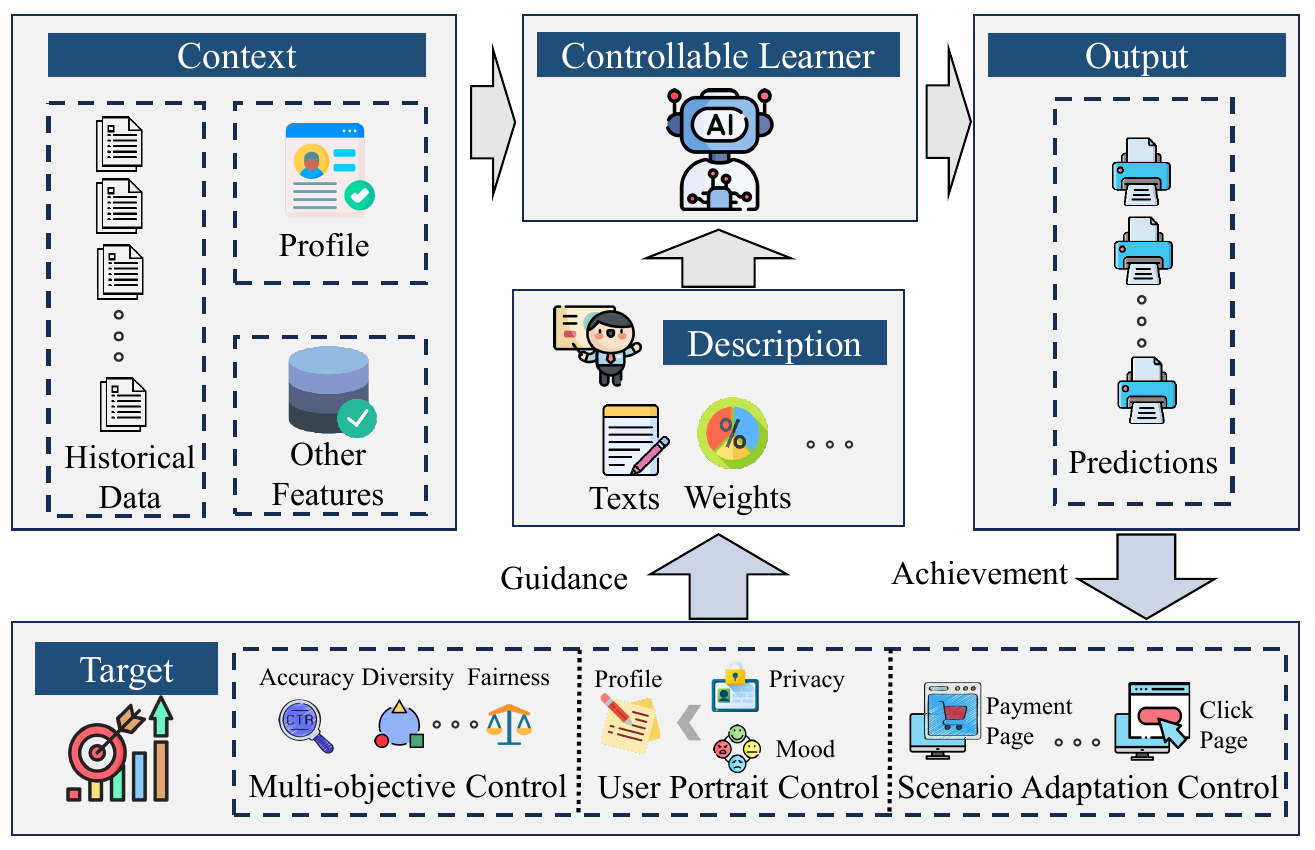}
    \caption{The procedure of CL in  Definition~\ref{def:CL:first}. Specifically, the context contains the historical interaction data, user profile, and other side information. The task target is categorized into three types: multi-objective control, user portrait control, and scenario adaptation control. The rationale behind this classification is detailed in Section~\ref{sec:why}. The task description is a representation of the task target, such as the preference weights of multi objective, natural language and so on.  The controllable learner consists of the control function $h$ and the learner $f$, taking the task description and context as input and outputting a new learner $f_{\mathcal{T}}$ to accomplish recommendation consistent of the task target.}
    \label{fig:CL:defination}
\end{figure*}

Under the definition, the procedure of CL could be shown as Figure~\ref{fig:CL:defination}. When the output space $\mathcal{Y}$ in Definition~\ref{def:CL:first} is not labels or lists but instead generated content, it is referred to as controllable generation.
In this survey, our primary focus lies on CL methods within information retrieval applications, which typically belong to controllable discrimination rather than controllable generation.

\subsection{Comparison with Other Topics}


Fairness, privacy, and interpretability are key aspects of trustworthy machine learning. Fairness ensures that model predictions do not exhibit bias against protected groups; privacy focuses on protecting sensitive user data from exposure; interpretability provides transparency into model behavior, enabling understanding and trust. Controllable learning (CL), by contrast, emphasizes the ability of a model to adapt to dynamic task requirements at test time without retraining, guided by a task requirement triplet consisting of task description, context, and target. While fairness and privacy are often enforced through training-time constraints, interpretability facilitates human understanding and is foundational for enabling test-time control. Within the CL framework, fairness can be formulated as a target in multi-objective control, privacy can be addressed via user portrait control by modifying context inputs, and interpretability can support interactive control through editable task descriptions in natural language or concept-based formats. Therefore, CL serves as a crucial mechanism for aligning model behavior with diverse and evolving user or platform objectives, bridging multiple dimensions of trustworthy machine learning.

%% file: sections/3_control_ir.tex
\section{Controllable Learning (CL) in Information Retrieval: Taxonomy}
\label{sec:taxonomy}
CL plays a crucial role in modern information retrieval (IR), where learning models must dynamically adapt to various task descriptions, such as users' specific information needs, without requiring extensive retraining. This capability ensures the delivery of personalized and relevant search results, thereby enhancing user satisfaction in IR systems.
Recent advancements underscore the importance of controllability in contemporary IR platforms~\cite{li2023breaking, wang2022user, mysore2023editable}. The following sections delve into the taxonomy of CL in IR, exploring who controls and what is controllable in IR systems. Additionally, we also present some representative works and provide a detailed description of their control methods in terms of ``what'' and ``who'' categories. (Table \ref{tab:summary_cl_methods}).
\subsection{Why This Taxonomy}
\label{sec:why}

%

The rationale behind the taxonomy is as follows. From a practical standpoint, it is natural to ask: what do we aim to control in IR? In essence, the purpose of control can be broadly categorized into three types. First, we control for shifting interests to quickly satisfy the diverse and evolving needs of both users and platforms—this is denoted as multi-objective control. Second, we control for privacy protection by allowing users to edit their personal profiles, referred to as user-portrait control. Third, we control for environmental adaptation, enabling models to efficiently adjust to varying scenarios through explicit scenario descriptions—this is termed scenario adaptation control. In addition to the question of ``what to control'', we also consider ``who controls''. Both users and platforms have control needs, but as initiators and recipients of IR systems respectively, their control intents diverge. Hence, we distinguish their roles and objectives accordingly in our taxonomy. Finally, we delve into the technical dimension to categorize and summarize common approaches to CL, which are captured by the axes of ``How'' (i.e., control techniques) in Section ~\ref{sec:where_to_control} and ``Where'' (i.e., the stage in the pipeline where control is applied) in  Section~\ref{sec:CL_technology}.

Importantly, these practical needs directly correspond to the formal components in our definition of CL.  We unify the CL process into a framework as illustrated in Figure 2, following our definition of CL. As shown in Definition~\ref{def:CL:first}, the execution of CL typically requires a clearly specified task requirement. We define the task requirement triplet as follows: $\mathcal{T} = { \bm{s}_{\mathrm{desc}}, \bm{s}_{\mathrm{ctx}}, \bm{s}_{\mathrm{tgt}} } \in \Gamma$, where $\bm{s}_{\mathrm{desc}} \in \mathcal{D}_{\mathrm{desc}}$ denotes the task description, $\bm{s}_{\mathrm{ctx}} \in \mathcal{D}_{\mathrm{ctx}}$ denotes the task-related context, and $\bm{s}_{\mathrm{tgt}} \in \mathcal{D}_{\mathrm{tgt}}$ denotes the task target.  For the task target $\bm{s}_{\mathrm{tgt}} \in \mathcal{T}$, we explore the question of ``control for what'' in the IR scenario, i.e., ``what'' the target $\bm{s}_{\mathrm{tgt}}$ could be. However, the two parties in an IR system—users and platforms—often focus on different aspects of $\bm{s}_{\mathrm{tgt}}$, leading to divergent task formulations (i.e., $\bm{s}_{\mathrm{des}}$). Thus, we further investigate ``who'' (user or platform) is responsible for specifying the task description under different target settings, as this sheds light on their respective controllability intentions. Finally, to support future research, we categorize common techniques for CL (denoted  as ``How'' ), and map their functional roles to the corresponding modules of the CL framework (denoted as ``Where'') to facilitate structured understanding and community advancement.



\subsection{What is Controllable}
\label{sec:what_to_control}
This section aims to clarify ``what is controllable'' for a controllable learner in information retrieval applications.  Specifically, we explore the question: ``What does the task target $\bm s_{\mathrm{tgt}}$  in  Definition~\ref{def:CL:first} of CL could be?'' 
The first type of task target is multi-objective control. In real-world business scenarios, both users and platforms experience shifts in interests, especially in multi-objective settings, where the emphasis on different objectives varies over time. Therefore, enabling the model to adapt to these changing objective requirements is particularly important. 
The second task target is user portrait control, where users can edit the context, thereby influencing the recommendation output. Some platforms provide preference questionnaires, allowing users to express personal preferences independently. Similarly, some platforms allow users to adjust their profiles and interaction history, achieving specific needs such as privacy, personalization and so on.
The third task target is scenario adaptation control. Real-world platforms typically contain multiple content pages, where users’ transition data across different pages can be valuable. Additionally, in the time dimension, user behavior on some platforms is often specifically related to time segments (considered as a special scenario). Explicitly leveraging scenario information to control the recommendation system is therefore essential.

Following this classification, Section~\ref{sec:multi-faced_retrieval_objectives} provides a detailed analysis of existing work on multi-objective control. Section~\ref{sec:historical_behavior_control} focuses on user portrait control, and Section~\ref{sec:controllable_environmental_adaptation} summarizes existing research on scenario adaptation control.

\subsubsection{Multi-Objective Control} 
\label{sec:multi-faced_retrieval_objectives}
\begin{figure}[t]
    \centering
    \includegraphics[width=0.48\textwidth]{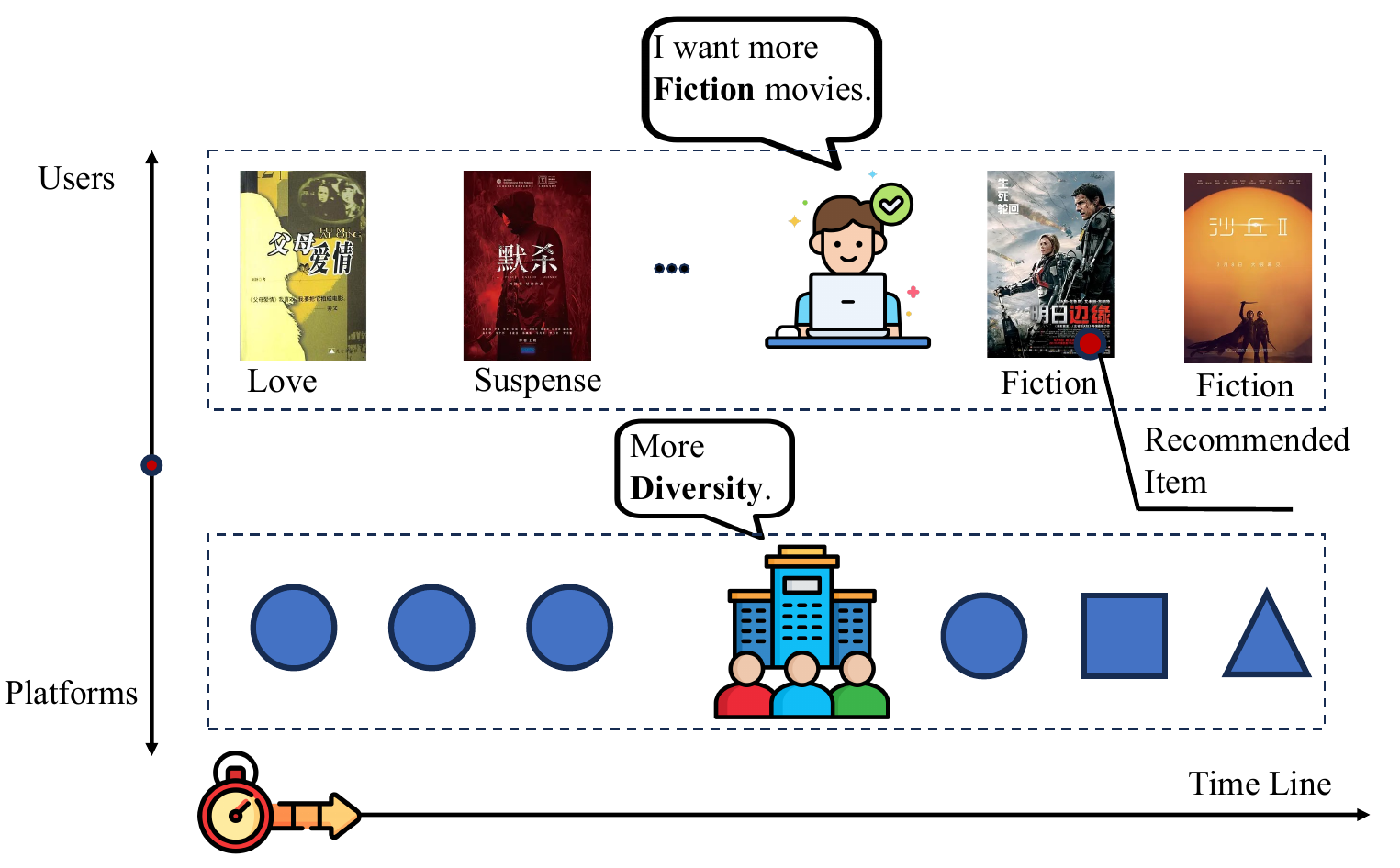}
    \caption{Illustration of the need for Multi-Objective Control. Users may have a temporary preference during the test stage (e.g., shifting from ``Love'' and ``Suspense'' to ``Fiction'' after a temporary description). The platform may instead focus more on the diversity of the outputs rather than accuracy during the test stage. This figure depicts the dynamic nature of shifting objectives for both the platform and users during test stage, underscoring the critical role of multi-objective control in recommender systems.}
    \label{fig:multiobjective}
\end{figure}

The preference (task targets) of the platform for multiple objectives (such as accuracy, diversity, and novelty) change over time. When facing a new task target, we expect a control function to directly map the learner $f$  to $f_{\mathcal{T}}$ that meets the demand based on the task description $\bm s_{\mathrm{{desc}}}$, where the $\bm s_{\mathrm{{desc}}}$ is defined as the preference for each objectives (in the explicit form of weights or natural language or the implicit form of interaction behaviors and so on). Moreover, at a finer granularity, user preferences for different categories can also be regarded as a form of multi-objective control. Figure~\ref{fig:multiobjective} illustrates an example of multi-objective control. Assuming the task target includes objectives related to accuracy and diversity, and the task description is represented as a two-dimensional vector indicating preferences for these objectives. From the platform's perspective, there is a strong emphasis on accuracy in a given period, leading to homogeneous recommendations (as shown by similar circles being recommended). Suppose the platform adjusts its business metrics to increase diversity in the output results, indicating a change in the task target. According to the definition of CL, we only need to modify the task description (i.e., increase the value of the diversity dimension in the two-dimensional vector). The control function can then directly map the learner to meet the new task target, instead of retraining the model based on the new task target.
On the user side, if we consider preferences for different types of items as different objectives, a similar analysis can be applied as in the platform-side case above.

ComiRec~\cite{cen2020controllable} strikes a balance between accuracy and diversity through hyperparameters in the aggregation module by capturing multiple user interests and integrating them into the recommendation process. UCRS~\cite{wang2022user} introduces a system where users can proactively reduce filter bubbles, controlling the recommendations they receive via hyperparameters that affect accuracy and coverage. Although the above two algorithms primarily focus on balancing multiple objectives, they also enable the possibility of dynamic control. For example, ComiRec introduces a controllable hyperparameter $\lambda$ in the multi-objective aggregation module to adjust diversity, and UCRS provides control coefficients $\alpha$ and $\beta$ to regulate accuracy, isolation, and diversity. These adjustments on hyperparameters do not trigger model retraining, thus providing a foundation for dynamic control during the test stage. The CMR framework~\cite{chen2023controllable} is specifically designed to address the dynamic nature of preferences across multiple objectives. In this framework, a hypernetwork takes dynamic preference weights as input to adjust the last few layers of the recommendation model, enabling dynamic control over objectives such as accuracy and diversity. These systems collectively focus on achieving a harmonious balance among the various objectives of recommendation quality, ensuring that users receive accurate, diverse, and relevant recommendations.  PadiRec~\cite{shen2024generating} treats multi-objective models under different preference weights as learning targets, considering the preference weights as conditions. Through conditional learning with a diffusion model, it captures the relationship between preference weights and the parameters of the multi-objective model. During the testing stage, by inputting the desired preference weights, the diffusion model can customize the model in real time according to these specifications, enabling control and test-time adaptation. RecLM-gen~\cite{lu2024aligning} utilize LLMs to follow recommendation-specific instructions. They utilize conventional recommendation models to distill recommendation capabilities into LLMs via labels. They then propose a reinforcement learning-based alignment procedure to enhance the generalization ability of LLMs, which includes category control and category proportion control.


\subsubsection{User Portrait Control} 
\label{sec:historical_behavior_control}

\begin{figure}[t]
    \centering
    \includegraphics[width=0.48\textwidth]{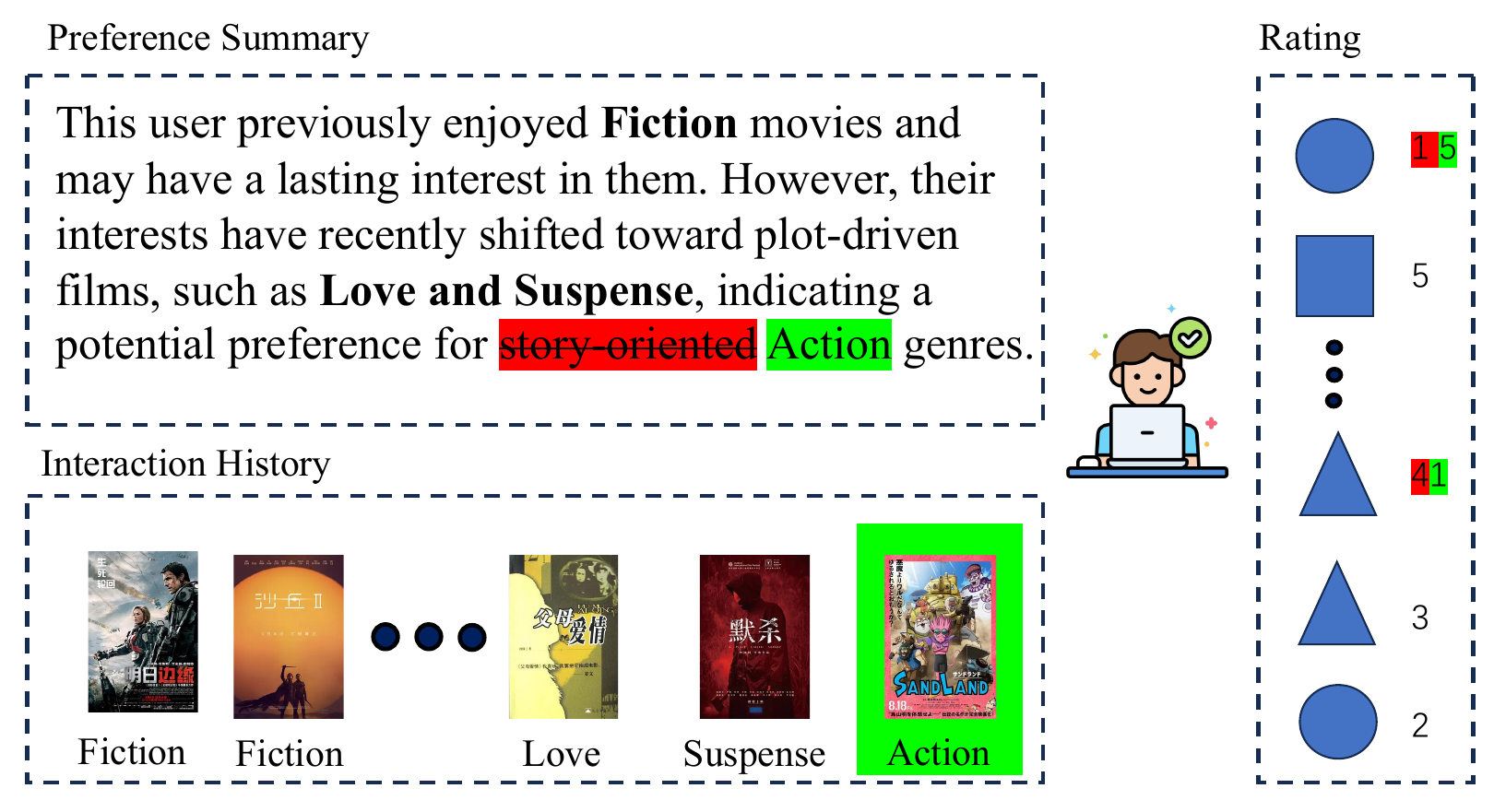}
    \caption{The Examples on User Portrait Control. The preference summary represents an aggregation of the user's interaction history. The list on the right shows the output items along with their ratings. Deletions are marked in red, and additions are marked in green. This figure illustrates three forms of user modification, demonstrating that user portrait control can be achieved by modifying and editing the interaction history, preference summary, and ratings.}
    \label{fig:userpotrait}
\end{figure}

On some platforms, the user profile or user history data can be edited to comply with privacy rules and provide users with greater control. Following our definition, this means that the context $\bm s_{\mathrm{ctx}}$ is editable, i.e., controllable, while in this case, the control function $h$ is a projection mapping that does not alter the learner $f$, specifically, $f = h(f, \bm s_\mathrm{desc}, \bm s_\mathrm{ctx})$. The task target is achieved by editing the input of the learner $f$. Some examples on the User Portrait Control are shown in Figure~\ref{fig:userpotrait}.

Some work in recommender systems involves allowing users to manage the context including their past interactions and profile information. LACE~\cite{mysore2023editable} allows users to edit their profiles by selecting or deselecting concepts, directly influencing the recommendations. IFRQE~\cite{wang2024would} lets users delete or modify past interactions to refine their recommendation experience.  TEARS~\cite{penaloza2024tears} summarizes user preferences in a text format, employing optimal transport to align summary-based embeddings with history-based black-box embeddings during the training stage. This allows users to edit their summaries, enabling them to control the recommendations they receive during the testing stage.
Together, these methods empower users to shape their digital footprint and ensure that recommendations better align with their evolving interests and preferences. CMBR~\cite{gou2024controllable} is designed for the specific scenario of skin recommendations in games, leveraging multi-domain user information and various types of user behavior data. During the training phase, CMBR inserts a special token before each type of behavior in the sequence, enabling the sequential model to learn the relationship between each special token and its corresponding behavior. In the test phase, by appending the desired special token at the end of the sequence, the model can predict the specific behavior associated with that token among the multiple behavior types.
UCR~\cite{tan2023user} proposes a user controllable recommendation framework with two essential properties. The first is retrospective controllability with retrospective explanation, which enables users to identify how their previous behaviors, such as clicks, purchases, and likes, contribute to current recommendations. The second is prospective controllability with prospective explanation, which offers users insight into how their interaction with the currently recommended items will influence future recommendations. Thus users can effectively control their future recommendations by operating in an informed manner.
\begin{figure}[t]
    \centering
    \includegraphics[width=0.48\textwidth]{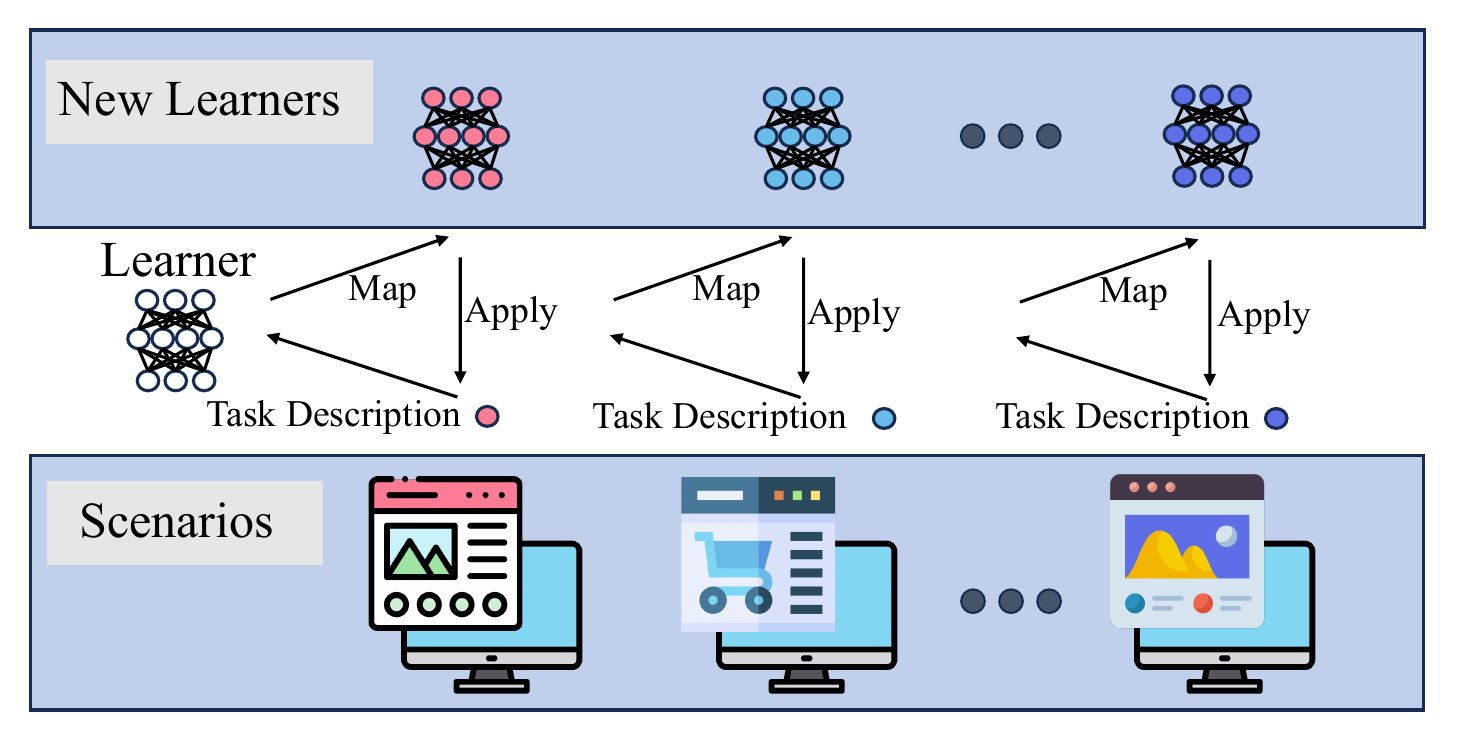}
    \caption{The Workflow of Scenario Adaptation Control. Unlike traditional settings where a learner is fixed to a specific scenario, Scenario Adaptation Control aims to address the challenge of learner adaptation under continuous scenario switching during the test stage. When a new scenario arises, the control function maps the general learner to a scenario-specific learner based on the corresponding task description. This allows the scenario-specific learner to be applied to the new context without the need for retraining.}
    \label{fig:scenarioadaptation}
\end{figure}

\subsubsection{Scenario Adaptation Control} 
\label{sec:controllable_environmental_adaptation}

In real-world recommendation systems, platforms often consist of multiple content pages, thereby creating multiple scenarios. From a temporal perspective, different time segments can also be viewed as distinct scenarios. A unified model helps fully leverage information across these scenarios, but it typically requires control to adapt the model to the current scenario. Following our definition, $\bm s_\mathrm{desc}$ needs to include scenario-specific side information. Thus, we can use the control function $h$ to map the learner $f$ to a scenario-specific learner $f_{\mathcal{T}}$. The work flow of the Scenario Adaptation Control is shown in Figure~\ref{fig:scenarioadaptation}. It is worth noting that some studies focus on transitions between multiple tasks, which we also consider as part of scenario adaptation.

Compared to dynamic online learning \cite{zhang2018adaptive, wan2021projection,zhao2023non}, controllable scenario adaptation is an explicit method that involves describing scenario-specific side information and inputting them into the controllable learner, rather than solely relying on the feedback of specific scenarios.  Hamur~\cite{li2023hamur} proposed a shared hypernetwork for dynamically generating the parameters of adapters to capture implicit information across different domains. The input to the hypernetwork consists of instance-level embeddings that carry domain-specific information, from which it generates adapter weight parameters. HyperBandit~\cite{shen2023hyperbandit} utilizes a hypernetwork to explicitly model the binary relationship between periodic external environments and user interests. In the testing stage, HyperBandit takes time features as input and dynamically adjusts the recommendation model parameters to account for time-varying user preferences. PEPNet~\cite{chang2023pepnet} introduces a gating mechanism that allows prior information to be injected into the model. This mechanism processes prior information with different personalized semantics through a two-layer neural network structure, generating personalized gating scores that adaptively control the importance of the prior information. DTRN~\cite{liu2023deep}  employing a hypernetwork to dynamically control task-behavior parameter interactions and refining representations with task-specific context awareness, DTRN effectively mitigates task interference and enhances representation specificity.

\subsection{Who Controls} 
\label{sec:who_controls}
\begin{figure}[t]
    \centering
    \includegraphics[width=0.48\textwidth]{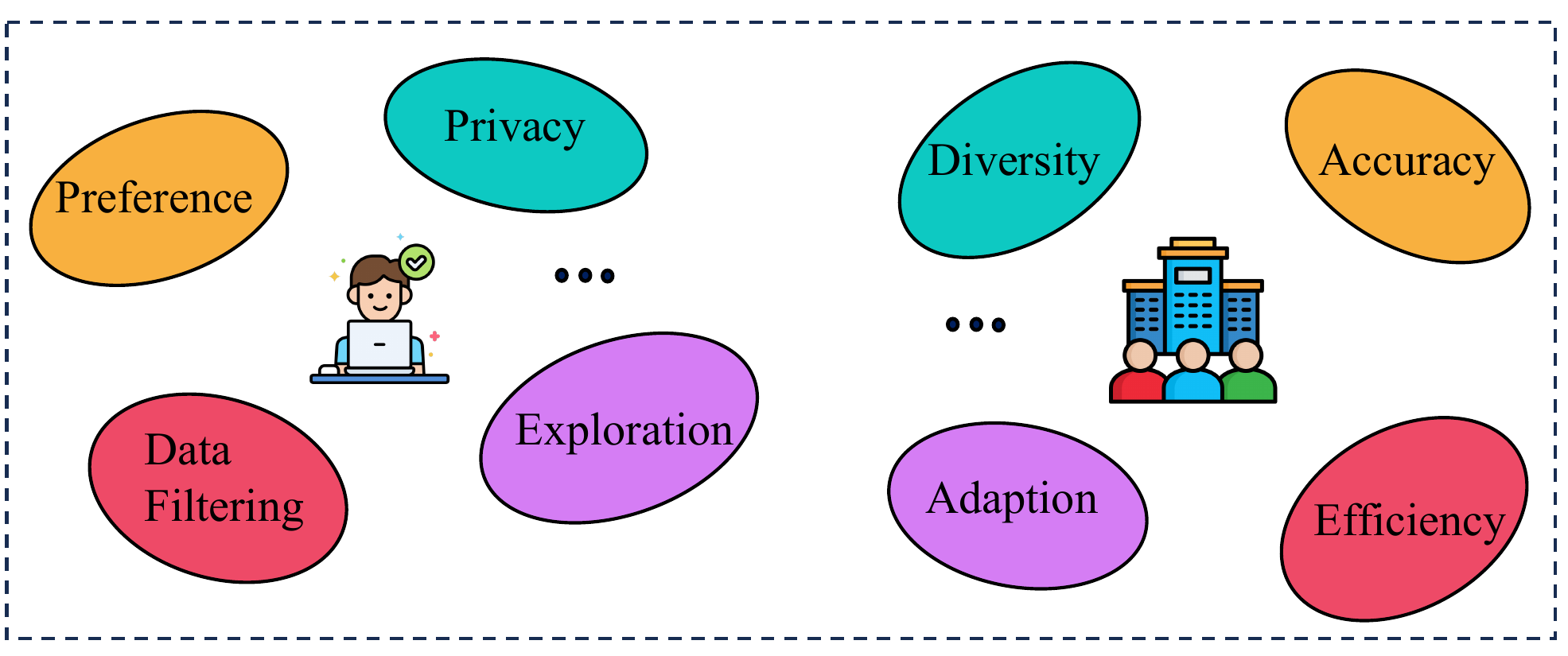}
    \caption{The Objects of User-Centric Control and Platform-Centric Control.}
    \label{fig:whocontrol}
\end{figure}

This section aims to clarify ``who controls'', which refers to who proposes the task description $\bm s_{\mathrm{desc}}$  in Definition~\ref{def:CL:first}.  
According to our research, controllability in recommendation systems can be categorized from the perspective of ``who controls'' into two aspects: user-centric and platform-centric. Users may have various needs, such as protecting their privacy, removing accidental interactions from their history (i.e., data filtering),  indirectly expressing preferences~\cite{jin2017different}, and explore the novel items. The platform may aim to increase the diversity of the learner’s output to promote the exposure of less popular items, adjust the trade-off between
multi objectives (e.g., accuracy and diversity), utilize the adaption between multi scenarios to promote the performance, and unify controllable model for all to cut the cost (i.e., efficiency). The objectives of each side are shown in Figure~\ref{fig:whocontrol}.

\subsubsection{User-Centric Control} 
\label{sec:user_centric_control}
User-Centric Control focuses on explicitly defining users' preferences for specific content. This means that users have a clear awareness of their own interests (i.e., task target) and provide this information to the recommendation system in a specified format (i.e., task description), such as through questionnaires, weighting buttons, or natural language. Additionally, users may not explicitly express their preferences. Instead, they may convey them through behaviors, such as interactions with recommendation results. 

\textbf{Explicit Control.} In some applications, users explicitly defining their preferences, which means that users have a clear awareness of their own interests (i.e., task target). Then the users clearly provide this information to the recommendation system in a specified format (i.e., task description), such as through questionnaires, weighting buttons, or natural language.

Supervised $\beta$-VAE~\cite{nema2021disentangling} proposes a supervised disentangled $\beta$-VAE-based recommendation model, which explicitly assigns latent dimensions to user preference attributes and aligns them with labeled features to achieve disentanglement. By incorporating classification loss for effective supervision on partially labeled data, the approach generates interpretable and controllable recommendations while maintaining high recommendation performance and enabling fine-grained user control over the output. LP~\cite{luo2020latent} presents a method allowing users to explicitly control recommendation results through iterative critiques. It extends a linear embedding-based recommendation approach by integrating subjective keyphrase feedback and modulating critique strength in a multistep process. This method reduces interactions required for satisfactory recommendations and improves success rates, showcasing enhanced user-centric recommendation adaptability.
TEARS~\cite{penaloza2024tears} captures user preferences in a text-based format and leverages optimal transport to align embeddings from summaries with black-box embeddings based on user history during training. This setup allows users to edit their summaries directly, giving them control over the recommendations they receive during testing. LangPTune~\cite{gao2024end}, a framework that enables explicit user control over recommendation systems through natural language profiles. By integrating end-to-end training with reinforcement learning and contrastive learning, it optimizes both recommendation accuracy and the interpretability of user preferences. This approach ensures users can directly modify their profiles to steer recommendations, enhancing transparency and personalization. GOMMIR~\cite{wu2023goal} incorporates user language as a reward signal into reinforcement learning optimization, enabling explicit user-side controllability through goal-oriented rewards and further optimizing recommendation performance.
Promptriever~\cite{weller2024promptriever} introduces the first retrieval model capable of being prompted like a language model. It not only achieves strong performance on standard retrieval tasks but also follows instructions, enabling explicit user control.

\textbf{Implicit Control.}  Users may not explicitly express their preferences. Instead, they may convey them through behaviors, such as interactions with recommendation results. Under this setting, the task description may not exist. The user's preference can be reflected through modifications to the context, which encompasses interaction history, general user profile and item description.

UCRS~\cite{wang2022user} proposes four control commands to reduce filter bubbles. Fine-grained user control allows users to increase items associated with their specific characteristics. Coarse-grained user control helps users reduce items tied to their own profile traits. Fine-grained item control enables adjustments like increasing items from specific categories. Coarse-grained item control allows reducing items from the most frequent category in the user's interaction history.  In practical scenarios, users may not want all their behaviors to be included in the model training process. IFRQE~\cite{wang2024would} lets users decide which of their interactions should contribute to the training of the recommendation. The models are then optimized to maximize utility, which considers the trade-off between recommendation performance and respecting user preferences. LACE~\cite{mysore2023editable} enables users to edit concept-based profiles for personalized text recommendations. LACE represents each user as a set of human-readable concepts by retrieving items the user has interacted with. These concept-based and editable user profiles are then used to make recommendations. The model’s design enables intuitive interaction with transparent user profiles to control recommendations. InstructAgent~\cite{xu2025instructagent} introduces a new ``User-Agent-Platform'' paradigm and proposes an instruction-aware agent, InstructAgent, which acts as an intermediary between users and the recommendation system. It addresses issues in traditional recommendation systems such as compromised user interests and insufficient personalization, thereby providing a new paradigm for user control.

However, whether explicit or implicit, these distinctions merely represent a form of user-controllable implementation. Their underlying purpose remains consistent: to foster interest, protect privacy, eliminate noise, and explore the unknown.


  
\subsubsection{Platform-Mediated Control} 
\label{sec:platform_mediated_control}

Platform-mediated control is characterized by algorithmic adjustments and policy-based constraints that a platform can impose on the recommendation process. In terms of form, control requirements are still expressed as $\bm s_{\mathrm{desc}}$. However, unlike user-centric control, platform-mediated control represented by $\bm s_{\mathrm{tgt}}$ focuses more on  the adjustment to various and dynamic tasks during the test time phase.

For instance, recent work aims to consider a general platform-side framework to control the algorithms in recommender systems and to avoid the creation of isolated communities and the filter bubble effect~\cite{li2023breaking}. 
Coincidentally, CCDF~\cite{zhang2024practical} controls the diversity of categories in recommendations by selecting top categories based on user-category matching and constrained item matching, aiming to alleviate echo chamber effects and broaden user interests. Introducing diversity alone may lead to a decline in recommendation performance, prompting the platform to focus on controlling multiple objectives. The CMR~\cite{chen2023controllable}, Padirec~\cite{shen2024generating} both utilize parameter generation for downstream models based on different objective weights, allowing the platform to adapt recommendations to various user groups or environmental changes at test time. In addition to diversity, user interests are also multifaceted. ComiRec~\cite{cen2020controllable} captures multiple interests from user behavior sequences and generates the top N items controllable by the platform in the aggregation module. Additionally, scenario signals and task signals are also utilized as control conditions to enhance the model's adaptability. SAMD~\cite{huan2023samd} offers a method for scenario-aware and model-agnostic knowledge dissemination across diverse contexts, enhancing the system's ability to tailor recommendations precisely.  HyperBandit~\cite{shen2023hyperbandit}  takes into account the correlation between time features and user preferences. This approach enables dynamic adaptation to evolving user preferences over time.
Additionally, DTRN~\cite{liu2023deep} obtains task-specific bottom representation explicitly by making each task have its own representation learning network in the bottom representation modeling stage. 

In summary, the platform-mediated control primarily aims to optimize recommendation strategies to meet specific objectives, such as increasing diversity or improving recommendation accuracy. Beyond these goals, there are more complex control tasks, such as control among multiple objectives and the transitions between scenarios and tasks. Overall, platform-mediated control focuses on equipping the model with the ability to adapt to specific task directives during the test time phase.

%% file: sections/4_methods.tex
\section{How Control is Implemented}
\label{sec:implement_control}
In Section~\ref{sec:taxonomy}, the existing works are categorized based on controllable objectives (i.e.,what) and controllers (i.e., who). In this section, we summarize the common control methods and their application positions within the recommendation system workflow, corresponding to Section~\ref{sec:CL_technology} and Section~\ref{sec:where_to_control}. 

\subsection{Controllable
Learning Techniques}
\label{sec:CL_technology}


We analyze and summarize the controllble learning (CL)  techniques for implementation of the control function $h(\cdot)$, where $h(\cdot)$ maps the learner $f$ to a new learner $f_{\mathcal{T}}$ based on the triplet $\mathcal{T}$ following the definition in Definition~\ref{def:CL:first}. The control function is regarded as a key mechanism for exerting control over the learner. However, to the best of our knowledge, there are several variations of this framework, which we will introduce in the following sections.

\subsubsection{Rule-Based Techniques}
\label{sec:rule_based_control}

Rule-based controllable techniques have proven to be indispensable in many scenarios such as recommender systems, information retrieval, and text generation due to its direct mode. Following Definition \ref{def:CL:first}, the $\bm s_{\mathrm{tgt}} \in \mathcal{D}_{\mathrm{tgt}}$ of the triplet typically represents the system performance we expect to achieve, such as security, fairness etc and the \( \bm s_{\mathrm{desc}} \in \mathcal{D}_{\mathrm{desc}} \) is always implicit. Rule-based methods acting as a patchwork solution, enhance system performance from a product perspective. Specifically, rule-based methods can be categorized into two types: pre-processing and post-processing. Formally, the control function $h$ is defined as a mapping:
$
h: f \rightarrow f \circ g_{\text{rule}}
$
in the context of pre-processing, whereas it is defined as:
$
h: f \rightarrow g_{\text{rule}} \circ f
$
in the context of post-processing, where $ f $ represents the base function and $g_{\text{rule}}$ denotes the rule-based control mechanism. For pre-processing algorithms, the strategy involves handling the context information including user profiles, interaction history, and others, denoted as $\bm s_{\mathrm{ctx}}$ in Definition~\ref{def:CL:first}, with the aim of altering the representations of users and items to meet specific requirements, such as privacy preservation or fairness. In contrast, post-processing algorithms, such as removing outdated items or promoting less popular items, can directly enhance diversity. For example, on an e-commerce platform, a user may not want the recommendation system to perceive their preference for a certain category of products (e.g., due to bias introduced by accidental clicks), which represent the task target $\bm s_{\mathrm{tgt}}$. A rule-based controllable recommendation system in the pre-processing stage should allow the user to specify a control mechanism $g\_{\text{rule}}$ to process the input to the recommendation system (i.e., the user's interaction history), filtering out unintended user behaviors. Thus, it avoids modeling biased information and improves recommendation accuracy.


Rule-based techniques involves applying predefined rules to the inputs or outputs of recommender systems to refine and enhance the data or results. In terms of privacy protection, the platform can achieve security and control through modifications and synthesis of input data~\cite{acs2018differentially,bindschaedler2017plausible, cunningham2021privacy}.
In the aspect of diversity, rule-based heuristic algorithms for recommended items based on user needs can also prove effective~\cite{carbonell1998use, kulesza2012determinantal}. Specifically, MMR~\cite{carbonell1998use} iteratively selects items that maximize a combined criterion of relevance to the user's query and minimal redundancy with already selected items, which ensures that the final recommendation list is both pertinent and varied.
Additionally, rule-based methods are also highly effective in personalization and fairness. Nandy et al.~\cite{nandy2022achieving} explore how fairness can be achieved through post-processing in web-scale recommender systems. By applying rule-based adjustments, the system can mitigate biases and ensure that recommendations are fair and equitable across different user groups. Similarly, Antikacioglu and Ravi ~\cite{antikacioglu2017post} focus on improving diversity in recommender systems through post-processing techniques. By incorporating rules that promote diversity, the system can offer a broader range of recommendations, preventing the issue of over-specialization and improving user satisfaction.
In addition, in the explainability of recommendations, rule-based methods have demonstrated a simple yet effective performance. Le et al.~\cite{le2023combining} combine embedding-based and semantic-based models for post-hoc explanations in recommender systems. Rule-based post-processing plays a crucial role in providing clear and understandable explanations for the recommendations, enhancing user trust and system transparency.

Rule-based techniques is powerful in making AI systems more controllable across various applications. It allows for the fine-tuning of AI outputs to meet specific requirements, ensuring accuracy, fairness, and reliability. As AI continues to evolve, the integration of rule-based post-processing methods will remain crucial in enhancing the controllability and overall performance of AI systems.

\begin{figure}[t]
    \centering
    \includegraphics[width=0.48\textwidth]{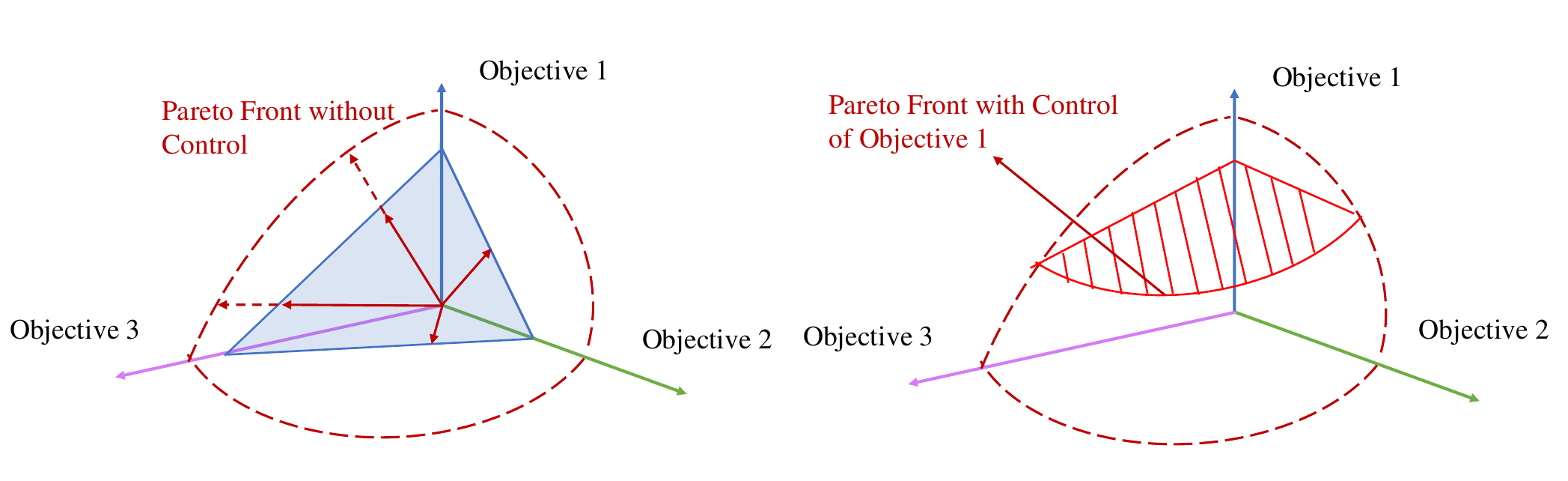}
    \caption{The illustration of controllable pareto optimization.}
    \label{fig:pareto}
\end{figure}

\subsubsection{Pareto Optimization}
\label{sec:pareto_optimization}
In the domain of machine learning, the concept of controllability has garnered significant attention due to the necessity to balance multiple, often conflicting objectives inherent in lots kinds of tasks. The integration of Pareto optimality principles from multi-objective optimization (MOO) has proven particularly instrumental in crafting algorithms that can navigate these trade-offs effectively. Following Definition \ref{def:CL:first}, the $\bm s_{\mathrm{tgt}}$ of the triplet typically represents a single  Pareto optimal solution that simultaneously satisfies multiple target objectives. The $\bm s_{\mathrm{desc}}$ generally consists of a set of multi-objective weights or constraints that enable the system to achieve the target. The $\bm s_{\mathrm{ctx}}$ usually depends on the scenario. In the field of information retrieval, it often includes candidate documents, user profiles, and other relevant context.
As illustrated in Figure~\ref{fig:pareto}, the key challenge of Pareto optimization in the domain of CL is determining how to guide the learner $ f $ to achieve Pareto optimality under given constraints.

Pareto MTL~\cite{lin2019pareto2} pioneers this field by proposing an algorithm that decomposes the MTL problem into a series of subproblems, each representing a unique trade-off preference. This approach yields a set of Pareto optimal solutions, allowing for flexibility in selecting the most appropriate solution based on specific task requirements. Furthermore, Controllable Pareto MTL~\cite{lin2020controllable} advances this concept by introducing a framework that enables real-time adjustments to these trade-offs, providing a more dynamic and adaptable system capable of responding to changing user preferences and task demands. In parallel, research on user-side personalized preferences has also emerged progressively. Ruchte et al.~\cite{ruchte2021scalable} propose a technique that conditions deep neural networks directly on preference vectors, thereby generating a well-distributed set of Pareto solutions. This work addresses the computational challenges associated with deep learning and sets the stage for more efficient and effective exploration of the Pareto front in complex MTL scenarios. A novel framework PHN-HVI~\cite{hoang2023improving} is proposed, which uses a hypernetwork to generate diverse solutions and improve the quality of the Pareto front in multi-objective optimization problems by optimizing the Hypervolume indicator, significantly outperforming existing methods. Similarly, PHNs~\cite{navon2020learning} utilizes a single hypernetwork approach to efficiently learn the entire Pareto front for multi-objective optimization problems, generalizes to unseen operating points, and enables post-training selection of optimal models based on runtime preferences. Complementing these efforts, research has also focused on the continuous exploration of Pareto sets, which provides a more nuanced understanding of the trade-offs in MTL. Ma et al.~\cite{ma2020efficient} introduce an innovative method that constructs continuous approximations of local Pareto sets, facilitating the applicability of MOO principles to the realm of deep MTL. Some researches have encompassed multi-objective optimization and Pareto efficiency in recommendation algorithms~\cite{xie2021personalized, lin2019pareto}. PAPERec~\cite{xie2021personalized} introduces personalized Pareto optimality in multi-objective recommendation achieving adaptive control of personalized objective weights. This model allows for dynamic balancing of conflicting objectives, such as accuracy and diversity, tailored to individual user preferences.  Lin et al.~\cite{lin2019pareto} introduces a general framework for reconciling objectives such as Gross Merchandise Volume (GMV) and Click-Through Rate (CTR) in E-Commerce Recommendation. 
MoFIR~\cite{ge2022toward} leverages multi-objective reinforcement learning to provide controllability over the trade-off between utility and fairness in recommendation systems. By introducing a conditional network and a Pareto-efficient optimization approach, the framework allows decision-makers to dynamically adjust recommendation outcomes based on predefined utility-fairness preferences. 

Most of the aforementioned studies aim to achieve customized Pareto optimality. However, they do not focus on test-time control, which means the learner $f$ still requires retraining to accommodate new user requirements and achieve Pareto optimality.  We believe that test-time control to achieve Pareto optimality without retraining is a promising research direction and anticipate increased attention to this area in the future.

\subsubsection{Hypernetwork}
\label{sec:parameter_control}

Hypernetworks~\cite{ha2016hypernetworks}, which are neural networks designed to generate the parameters for another network, offer a flexible and efficient way to manage and adapt model parameters dynamically. 
Therefore, they have gradually emerged as a key technique for enhancing model controllability due to their explicit control capabilities. 
Following Definition \ref{def:CL:first}, the $\bm s_{\mathrm{desc}}$ typically represents the description of the task or domain, usually serving as the input for the hypernetwork. The hypernetwork act as the control function $h$ taking $\bm s_{\mathrm{desc}}$ as input, mapping the learner through parameter generation. Thus the mapped learner could achieve the corresponding $\bm s_{\mathrm{tgt}}$. Take the streaming recommendation as an example. It is known that data undergoes distributional shifts over time (e.g., users may prefer anime on weekends and educational videos during weekdays). Traditional train-test two-stage models fail to generalize to out-of-distribution data at test time. A feasible solution is to use a hypernetwork as the control function $h$, taking the time feature as the task description $\bm{s}_{\mathrm{desc}}$. By feeding $\bm{s}\_{\mathrm{desc}}$ into the hypernetwork, it outputs all or part of the weights for the learner $f$, thereby enabling the customization of a specific model for different time periods during the test stage and allowing the model to adapt to distributional shifts in the data.

Galanti and Wolf~\cite{galanti2020modularity} discuss the modularity of hypernetworks, emphasizing their ability to decompose a complex task into smaller, manageable sub-tasks. This modular approach allows for the generation of highly specialized parameters tailored to specific retrieval tasks, thereby improving accuracy and efficiency. Hyperprompt~\cite{he2022hyperprompt} introduces a prompt-based task-conditioning method for transformers that leverages hypernetworks. This approach dynamically generates prompts that condition the transformer model to adapt to different retrieval tasks, enabling more precise and context-aware information retrieval. Hamur~\cite{li2023hamur} propose a hyper adapter designed for multi-domain recommendation. Hamur utilizes hypernetworks to adaptively generate domain-specific parameters, thereby enhancing the model's ability to recommend items across different domains with high relevance and accuracy. Furthermore, CMR~\cite{chen2023controllable} explore the use of policy hypernetworks for controllable multi-objective re-ranking in recommender systems. By leveraging hypernetworks to generate parameters that balance multiple objectives, such as relevance and diversity, the system can dynamically adjust its recommendations based on user preferences and system goals.
Similarly, HyperBandit ~\cite{shen2023hyperbandit} maintains a neural network capable of generating the parameters for estimating time-varying rewards, taking into account the correlation between time features and user preferences. This approach enables dynamic adaptation to evolving user preferences over time.
Mahabadi et al. \cite{mahabadi2021parameter} discuss parameter-efficient multi-task fine-tuning for transformers via shared hypernetworks, allowing for the efficient sharing of parameters across multiple tasks, reducing the overall computational load and improving the scalability of transformer models. Hypencoder~\cite{killingback2025hypencoder} employs a hypernetwork to generate a dedicated lightweight neural network for each query, which is then used to score documents. This approach overcomes the representational limitations of traditional vector dot product methods and enables fine-grained control over the relevance evaluation function.

In conclusion, hypernetworks represent a powerful tool for parameter control in information retrieval and recommender systems. By enabling dynamic and task-specific parameter generation, hypernetworks can significantly improve the performance and adaptability of AI models, paving the way for more sophisticated and responsive systems.

\subsubsection{Other Methods}
\label{sec:other_methods}
There are also some control methods in machine learning, which is not strictly belong to the above three classifications.  Some studies~\cite{bhargav2021controllable, luo2020latent, nema2021disentangling, wang2021controllable} propose the concept of \textbf{Disentanglement} , which decouples user interests into specific dimensions within the latent space to facilitate control. Specifically, Multi-VAE and Supervised $\beta$-VAE~\cite{bhargav2021controllable,nema2021disentangling}  provide the `knobs' for users, with each knob corresponding to an item aspect. The latent space where generative factors (here, a preference towards an item category like genre) are captured independently in their respective dimensions, thereby enabling predictable manipulations. LP~\cite{luo2020latent} models user-item interactions and user requirements in a linear space, enabling controllable user updates by integrating user requirements into user features through a weighted summation approach. 
CGIR~\cite{wang2021controllable} propose a weakly-supervised method that can learn a disentangled item representation from user-item interaction data and ground the semantic meaning of attributes to dimensions of the item representation. During inference, CGIR start from the reference item and ``walk'' along the direction of the modification in the item representation space to retrieve a sequence of items in a gradient manner. \textbf{Reinforcement Learning} (RL) is a kind of classic method to achieve controllability.  Since these algorithms learn from the interaction with the environment (actually the reward function of the environment), it is possible for us to control the algorithms through dedicatedly designing the reward function. 
LangPTune~\cite{gao2024end} leverages reinforcement learning to enable the encoder to better encode implicit user interaction sequences into interpretable text. Similarly, Lu et al.~\cite{lu2024aligning} adopt a reinforcement learning approach to design fine-grained reward functions for control rules, allowing the model to accurately respond to specific control instructions. Qin et al.~\cite{qin2024enhancing} leveraged the reflective reasoning capabilities of LLMs to summarize user preferences from multiple perspectives and integrated reinforcement learning to dynamically adapt to changes in user preferences. 
Zhang et al.~\cite{zhang2024reinforcing} propose the user-oriented exploration policy, which controls exploration patterns across user groups by customizing intensities via a distributional critic, quantile-targeting actors, and regularization for diversity and stability
Additionally,
inspired by the exceptional general intelligence of Large Language Models (LLMs)~\cite{zhao2023survey}, researchers have begun to explore the LLMs application in recommender systems and other fields. GOMMIR~\cite{wu2023goal} enhances user control over the recommendation process by integrating REINFORCE with both verbal and non-verbal user feedback. It further optimizes recommendation performance through goal-oriented rewards and hard negative sampling. Similarly, MocDT~\cite{gao2025future} leverages a Decision Transformer to represent and prioritize multiple objectives by introducing adjustable multi-objective control signals. This enables the generation of recommendation sequences tailored to different goals without the need for retraining, thereby enhancing the flexibility and controllability of recommendation systems in multi-objective scenarios. RecLM-gen~\cite{lu2024aligning} constructs \textbf{Supervised Fine-Tuning} (SFT) tasks, augmented with labels derived from a conventional recommender model, explicitly improving LLMs' proficiency in adhering to recommendation specific instruction which could be regarded as task description $\bm s_{\mathrm{desc}}$ as defined in Sec.\ref{sec:formulation_of_CL}.
For example, DLCRec~\cite{chen2024dlcrec} is a novel framework designed to enhance diversity control in LLM-based recommender systems. By decomposing the recommendation task into three supervised fine-tuning tasks-genre prediction, genre filling, and item prediction—it enables fine-grained diversity control guided by user-defined signals. Apart from the aforementioned technical approaches, recent works have leveraged large language models (LLMs) to enable interpretable and controllable \textbf{Natural Language Interest Editing} (NLIE). Notable examples include LACE~\cite{mysore2023editable}, TEARS~\cite{penaloza2024tears}, and LangPTune~\cite{gao2024end}, among others.
\textbf{Test-Time Adaptation} refers to algorithms that do not modify the model through retraining; instead, they allow a pre-trained model to leverage unlabeled data during the testing phase before making predictions. This type of algorithm is well-suited for online scenarios, where time constraints are stringent. Some studies focus on adapting the model directly during the test stage to cope with dynamic environments. $T^2ARec$~\cite{zhang2025test} innovatively integrates Test-Time Training into sequential recommendation by introducing two alignment-based self-supervised losses.
HyperBandit~\cite{shen2023hyperbandit} employs a hypernetwork to dynamically adjust the model based on temporal features of the test data before making predictions, without the need for retraining. Hamur~\cite{li2023hamur} employs a domain-shared hypernetwork that implicitly captures shared information across domains and dynamically generates adapter parameters to adapt to domain shifts during the test stage. Notably, not all hypernetwork-based algorithms fall under this category. This is because hypernetworks are typically designed to support proactive control by users or platforms, whereas test-time adaptation generally requires reactive adjustments based on test data.

Today, an increasing number of techniques are being applied to the field of controllable learning, and we look forward to the emergence of more novel and effective algorithms in the future.




\begin{table*}[t]
\centering
\caption{Summary of representative controllable learning (CL) methods for information retrieval. \textbf{``Who''} denotes the entity responsible for control (see Section~\ref{sec:who_controls}), including user-centric control and platform-mediated control.
\textbf{``What''} refers to what is task target (see Section~\ref{sec:what_to_control}) including, multi-objective control, user portrait control, scenario adaptation control.
\textbf{``CL Tech.''} describes the techniques for implementing control (see Section~\ref{sec:CL_technology}) including rule-based, pareto optimization, hypernetwork and others.
\textbf{``Where''} indicates the position where control is applied in the testing stage (see Section~\ref{sec:where_to_control}) including pre-processing methods, in-processing methods, and post-processing methods.}
\label{tab:summary_cl_methods}
\scriptsize
\renewcommand{\arraystretch}{1.0}
\begin{tabular}{rcrrrr}
\toprule
\multicolumn{2}{c}{Method Information} & 
\multicolumn{4}{c}{Paradigm of Controllable Learning} \\
\cmidrule(l){1-2} \cmidrule(l){3-6}
Method & Year & What & Who & CL Tech. & Where\\
\midrule

MocDT~\cite{gao2025future} & 2025 & multi-objective control & user-centric control& RL& in-processing\\
\addlinespace
PadiRec~\cite{shen2024generating}                   & 2024 & multi-objective control     & platform-mediated control         & hypernetwork                          & in-processing \\
\addlinespace
FollowIR~\cite{weller2024followir}& 2024& user portrait control &user-centric control& SFT & pre-processing \\
\addlinespace
InstructIR~\cite{oh2024instructir}& 2024& user portrait control &user-centric control& SFT & pre-processing \\
\addlinespace
RecLM-gen~\cite{lu2024aligning}            & 2024 & multi-objective control     & platform-mediated control         & SFT, RL                      & in-processing               \\
\addlinespace
IFRQE~\cite{wang2024would}                          & 2024 & user portrait control      & user-centric control  & others                         & pre-processing              \\
\addlinespace
TEARS~\cite{penaloza2024tears}             & 2024 & user portrait control      & user-centric control  & RL, NLIE                          & in-processing               \\
\addlinespace
CMBR~\cite{gou2024controllable}                     & 2024 & user portrait control      & user-centric control                    & SFT, RL                          & in-processing               \\
\addlinespace
LangPTune~\cite{gao2024end}                         & 2024 & user portrait control                          & user-centric control  & RL, NLIE                     & in-processing               \\
\addlinespace
CCDF~\cite{zhang2024practical}                      & 2024 & multi-objective control                          & platform-mediated control         & others                    & in-processing       \\
\addlinespace
CMR~\cite{chen2023controllable}           & 2023 & multi-objective control     & platform-mediated control         & hypernetwork                    & in-processing       \\
\addlinespace
LACE~\cite{mysore2023editable}            & 2023 & user portrait control      & user-centric control  & NLIE                      & pre-processing      \\
\addlinespace
UCR~\cite{tan2023user}                              & 2023 & user portrait control      & user-centric control                   & others                         & pre-processing               \\
\addlinespace
Hamur~\cite{li2023hamur}                            & 2023 & scenario adaptation control & platform-mediated control         & hypernetwork, TTA               & in-processing              \\
\addlinespace
HyperBandit~\cite{shen2023hyperbandit}              & 2023 & scenario adaptation control & platform-mediated control         & hypernetwork, TTA                  & in-processing       \\
\addlinespace
PEPNet~\cite{chang2023pepnet}                       & 2023 & scenario adaptation control & user-centric control                  & hypernetwork                          & in-processing               \\

\addlinespace
SAMD~\cite{huan2023samd}                            & 2023 &    scenario adaptation control                       & platform-mediated control         & hypernetwork                         & in-processing              \\
\addlinespace
DTRN~\cite{liu2023deep}                             & 2023 &  scenario adaptation control                        & user-centric control         & hypernetwork                          & in-processing               \\
\addlinespace
MoFIR~\cite{ge2022toward}&2022&multi-objective control&user-centric control & pareto optimization&in-processing\\

\addlinespace
UCRS~\cite{wang2022user}                            & 2022 & multi-objective control     & user-centric control  & others                          & pre-processing      \\
\addlinespace
PAPERec~\cite{xie2021personalized} 	&2021	& multi-objective control	&user-centric control	& pareto optimization &	in-processing\\

\addlinespace
Supervised $\beta$-VAE~\cite{nema2021disentangling} & 2021 & user portrait control                           & user-centric control  & Disentanglement                      & in-processing               \\
\addlinespace
ComiRec~\cite{cen2020controllable}                  & 2020 & multi-objective control     & platform-mediated control         & others                          & post-processing     \\
\addlinespace
LP~\cite{luo2020latent}                             & 2020 &  user portrait control                       & user-centric control  & Disentanglement                      &  in-processing              \\
\addlinespace
MMR~\cite{carbonell1998use}                         & 1998 &  multi-objective control       &    platform-mediated control  & rule-based                   & post-processing    \\
\bottomrule
\end{tabular}

\end{table*}

\subsection{Where to Control in Information Retrieval Models}
\label{sec:where_to_control} 
In this section, we categorize control learning methods into three distinct categories based on the inference process during the testing phase. Specifically, we classify them based on the stage at which the control function is applied: methods that adjust model inputs before inference are termed \emph{pre-processing methods}, those that operate on model parameters during inference are termed \emph{in-processing methods}, and those that act on model outputs after inference are termed \emph{post-processing methods}.

\subsubsection{Pre-Processing Methods} 
\label{sec:pre_processing_methods}
Pre-processing methods achieve the task target solely by transforming the model inputs without adjusting the model itself. Transformation of model inputs includes directly concatenating the task description onto the original feature vector. These methods are akin to in-context learning methods in large language models, where the task description serves as the prompt.

Mysore et al.~\cite{mysore2023editable} introduce the LACE model, which crafts user profiles using intelligible concepts extracted from user-document interactions. This innovation permits users to modify their profiles, thereby directly influencing the recommendation outcomes. Through comprehensive offline assessments and user studies, LACE's capacity to refine recommendation quality via user interaction has been convincingly validated. Lastly, Wang et al.~\cite{wang2024would} tackle the pivotal concerns of user privacy and the degree of control over data utilization in model training. This framework grants users the discretion to decide which of their interactions may be harnessed for training purposes, delicately balancing between optimizing recommendation effectiveness and honoring user preferences. The implementation of an influence function-based model, alongside an augmented model featuring multiple anchor actions, evidences the feasibility of reconciling high-quality recommendations with user consent.
UCRS~\cite{wang2022user} imagines a counterfactual world where out-of-date user representations are discarded, and estimates their effects as the difference between factual and counterfactual worlds. After deducting such effects, incorporates the control command into recommender inference. As to user-feature controls, it revises the user feature specified by the control command (e.g., changing age from middle age to teenager) to conduct the final inference at the two levels. As to item-feature controls, UCRS adopts a user-controllable ranking policy to control the recommendations w.r.t. item category. 
It is worth noting that, besides those works on recommender systems, the use of instructions in information retrieval is a relatively new development. TART~\cite{asai2022task} and Instructor~\cite{su2022one} adopt simple task prefixes during training. Meanwhile, several benchmark efforts explicitly focus on evaluating the instruction-following ability of retrievers, such as FollowIR~\cite{weller2024followir} and InstructIR~\cite{oh2024instructir}.


\subsubsection{In-Processing Methods} 
\label{sec:in_processing_methods}
In-processing methods adaptively adjust the parameters or, more loosely speaking, the hidden states of models upon receiving the task description and context to achieve the task target.
The CMR framework~\cite{chen2023controllable} harnesses the input preference vector to guide the hypernetwork, which in turn generates network parameters tailored to the desired balance of objectives, thus achieving the desired control during the testing stage. HyperBandit~\cite{shen2023hyperbandit} utilizes the periodic time information to inject to a hypernetwork, modeling the relationship between user preference with corresponding time block, achieving efficient user preference adaptation during testing stage. Similarly, the CCDF~\cite{zhang2024practical} employs a hyperparameter, denoted as `$k$', to directly manipulate the number of categories presented in the top-$k$ recommendations, allowing for precise control over the diversity of the recommended content. These methods demonstrate the utility of in-processing strategies in enhancing the adaptability and performance of recommender systems. 


\subsubsection{Post-Processing Methods}
\label{sec:post_processing_methods}
Post-processing methods, including reranking and result diversification, are crucial for refining the output of recommender systems. ComiRec~\cite{cen2020controllable} leverages a final aggregation module that balances dual objectives—accuracy and diversity—through a weighted summation approach to determine the top-k recommendations. This method ensures that the final selection of recommendations is not only precise but also diverse, catering to a broader range of user interests.
Some works achieve controllability by re-ranking.
MMR~\cite{carbonell1998use} is a post-processing technique in recommender systems that aims to balance relevance and diversity in the recommended items. By iteratively selecting items that maximize a combined criterion of relevance to the user's query and minimal redundancy with already selected items, it ensures that the final recommendation list is both pertinent and varied.



%% file: sections/5_resources.tex
\section{Evaluation for Controllable Learning in Information Retrieval Applications}
\label{sec:evaluation}
As far as we know, controllable learning was first proposed and defined by us, although it has been implicitly used in many works. However, existing works still lack specific evaluation criteria for controllable learning, and datasets to verify controllability have  also not been explicitly proposed. In this section, we provide some common metrics in IR and clarify that appropriate use of these metrics can verify controllability. We also present common datasets in IR.


\subsection{Metrics}
\label{sec:metric}

In the definition of controllable learning that we discussed, we expect the control function $h$ to output a new learner $f_{\mathcal{T}}$ that meets the task requirements $\mathcal{T}$. Therefore, in the evaluation phase, we need to assess whether $h$ can effectively control the output $f_{\mathcal{T}}$, specifically whether the performance of $f$ meets the task requirements. For example, using a parameter $\alpha$ to represent the degree of control over the performance $s$ of the output $f$ (such as NDCG, diversity, MAP, etc.), a simple case would be an approximately linear relationship between $\alpha$ and $s$~\cite{cen2020controllable},~\cite{chen2023controllable}. A straightforward idea is to calculate the correlation coefficient between $\alpha$ and $s$, such as the Pearson Correlation Coefficient~\cite{cohen2009pearson
} and the Spearman Rank Correlation Coefficient~\cite{zar2005spearman}, to measure the control effectiveness of the control function $h$.  

In specific fields like information retrieval, we can utilize the combined variation of multiple single-objective metrics to assess whether the performance of $f$ meets the task requirements. Aiming that, multi-objective optimization metrics are also necessary to assess that variation. In this section, we therefore introduce some common single-objective metrics used in the field of information retrieval and some common metrics on multi-objective optimization.

\subsubsection{Single-Objective Metrics} 
\label{sec:Single-objective metrics}
The objectives pursued by users and platforms can vary significantly. Users tend to prioritize the accuracy and novelty of recommendations, while platforms place greater emphasis on long-term revenue, recommendation efficiency, and other operational goals. In this chapter, we present a range of commonly used metrics for single-objective recommendation, categorized from various perspectives.

\emph{Accuracy.}
\vspace{0.3em}
Accuracy in recommender systems refers to how well the recommended items match the actual preferences or interests of users. It measures the system's ability to provide relevant items that the user is likely to engage with, such as clicking, purchasing, or interacting. Common metrics include:

\textbf{NDCG}~\cite{jarvelin2002cumulated} evaluates the effectiveness of information retrieval systems by accounting for both the relevance and rank of retrieved documents. It applies a logarithmic discount to lower-ranked items and normalizes against an ideal ranking for consistency. This makes NDCG essential for comparing search engine and recommender system performance across different queries: 
\begin{equation*}
\begin{aligned}
    \mathrm{NDCG}@k &= \frac{1}{N}\sum_{i=1}^{N}\frac{\mathrm{DCG}_i@k}{\mathrm{IDCG}_i@k},\\
    \mathrm{DCG}_i@k &= \sum_{j=1}^k\frac{2^{y_{ij}}-1}{\log_2(j+1)}, \\
\end{aligned}
\end{equation*}
where $N$ is the number of test samples, $y_{ij} \in \{0,1\}$ denotes the label of the $j$-th item. 

\textbf{Precision} is a metric used to evaluate the accuracy of information retrieval systems by measuring the proportion of retrieved documents that are relevant. It is calculated as the number of relevant documents retrieved divided by the total number of documents retrieved. High precision indicates that the system retrieves mostly relevant items, minimizing the presence of irrelevant information. This metric is particularly important in scenarios where presenting irrelevant items can negatively impact user experience, such as in search engines, recommender systems, and spam filtering:
\begin{equation*}
\mathrm{Precision}@k = \sum_{i=1}^{N} \frac{\widehat{L}^k_i \cap L^k_i}{\widehat{L}^k_i},
\end{equation*}
where $\widehat{L}^k_i$ is the top-$k$ list outputted by the IR model, $L^k_i$ is the ground-truth top-$k$ list.

\textbf{Recall} measures the effectiveness of information retrieval systems by determining the proportion of relevant documents that have been successfully retrieved from the total relevant documents available. It is calculated as the number of relevant documents retrieved divided by the total number of relevant documents. High recall indicates that the system retrieves most of the relevant items, which is essential in tasks where missing relevant information can have significant consequences, such as medical research or legal discovery: 
\begin{equation*}
    \mathrm{Recall}@k = \sum_{i=1}^{N} \frac{\widehat{L}^k_i \cap L^k_i}{L^k_i}.
\end{equation*}

\textbf{Hit Rate} evaluates the performance of recommender systems by assessing the presence of relevant items within the top-N recommendations provided to users. It is calculated as the number of users for whom at least one relevant item is included in the top-N recommendations, divided by the total number of users. Hit Rate is particularly useful for understanding the effectiveness of recommendation algorithms in scenarios where presenting at least one relevant option can significantly impact user satisfaction, such as in e-commerce or content streaming platforms:
\begin{equation*}
    \mathrm{Hit}@k = \sum_{i=1}^{N} \mathbb{I}(\widehat{L}^k_i \cap L^k_i \neq \emptyset),
\end{equation*}
where $\mathbb{I}(\cdot)$ is an indicator function.

\emph{Diversity.}
\vspace{0.3em}
Diversity in recommendation systems refers to the variety of items recommended to users. A diverse recommendation list includes items that are different from each other, avoiding redundancy and ensuring a broad range of options. Common metrics include:

\textbf{$\alpha$-NDCG}~\cite{clarke2008novelty} extends the NDCG metric to evaluate the diversity in information retrieval systems. By incorporating a parameter $\alpha$, it penalizes redundancy and rewards the retrieval of diverse, relevant documents. This makes $\alpha$-NDCG particularly useful for tasks like web search and recommender systems, where presenting varied content is crucial:
\begin{equation*}
\begin{aligned}
\alpha\text{-}\mathrm{NDCG}@k &= \frac{1}{N}\sum_{i=1}^{N}\frac{\alpha\text{-}\mathrm{DCG}_i@k}{\alpha\text{-}\mathrm{IDCG}_i@k},\\
\alpha\text{-}\mathrm{DCG}_i@k &= \sum_{j=1}^k \sum_{l=1}^{m} \frac{ t_{j,l}(1-\alpha)^{c_{j,l}}}{\log_2(j+1)}, \\
\end{aligned}
\end{equation*}
where $m$ is the number of subtopic. $t_{j,l}=1$ if the $j$-th item covers subtopic $l$ and $t_{j,l}=0$ otherwise. $c_{j,l}$ is the number of times the subtopic $l$ being covered by items prior to the $j$-th item.

\textbf{ERR-IA}~\cite{yan2021diversification} evaluates the effectiveness of information retrieval systems by considering user intent in the evaluation process. ERR-IA extends the Expected Reciprocal Rank (ERR) metric by incorporating a probabilistic approach to user intents, ensuring that the evaluation reflects the variety of user needs. It models user satisfaction as a function of the relevance of retrieved documents and their alignment with multiple user intents, making ERR-IA particularly useful for web search and other scenarios where understanding and catering to diverse user intents is crucial:
\begin{equation*}
\mathrm{ERR}\text{-}\mathrm{IA} = \frac{1}{N}\sum_{i=1}^{N} \sum_{j=1}^{k} \frac{1}{j} \sum_{l=1}^{m} \frac{1}{m} \frac{t_{il}}{2^{c_{j,l}} + 1}.
\end{equation*}

\textbf{Coverage} evaluates the comprehensiveness of information retrieval systems by determining the proportion of relevant items retrieved out of the total relevant items available across different queries or datasets. It is calculated as the number of unique relevant items retrieved divided by the total number of unique relevant items. High coverage indicates that the system retrieves a broad set of relevant items, making this metric essential for applications where capturing a wide range of relevant information is important, such as in comprehensive research databases, recommender systems, and digital libraries.

\begin{equation*}
    \mathrm{Coverage}@k = \frac{\cup_{i=1}^{N} \widehat{L}^k_i}{|\mathcal{I}|},
\end{equation*}
where $\mathcal{I}$ is the set of all items.


\emph{Fairness.}
\vspace{0.3em}
Fairness~\cite{li2022fairness,wang2023survey} in recommender systems aims to ensure that all users, regardless of demographic or social group, are treated equitably in the recommendations they receive. Group fairness is a subfield that focuses on ensuring that different groups (e.g., gender, race, age) are not disadvantaged by the recommendation process. We listed several commonly used group fairness metrics.

\textbf{Demographic Parity (DP)} ensures that different groups receive similar rates of 
recommendations, regardless of their past interactions or preferences:
\begin{equation*}
    \mathrm{DP} = \left|\frac{\sum_{i=1}^{|\mathcal{S}_0|} \hat{y}_i}{|\mathcal{S}_0|} - \frac{\sum_{i=1}^{|\mathcal{S}_1|} \hat{y}_i}{|\mathcal{S}_1|} \right|,
\end{equation*}
where $\mathcal{S}_0$ and $\mathcal{S}_1$ represent different groups divided based on sensitive features such as gender, $\hat{y}_i$ denotes the score predicted by the recommendation model.

\textbf{Equal Opportunity (EO)} ensures that groups with same feedback (e.g., clicks, purchases) are equally likely to receive relevant recommendations:
\begin{equation*}
    \mathrm{EO} = \sum_{y \in \{0,1\}} \left|\frac{\sum_{i=1}^{|\mathcal{S}_0^{y}|} \hat{y}_i}{|\mathcal{S}_0^{y}|} - \frac{\sum_{i=1}^{|\mathcal{S}_1^{y}|} \hat{y}_i}{|\mathcal{S}_1^{y}|} \right|,
\end{equation*}
where $\mathcal{S}_0^{y}$ and $\mathcal{S}_1^{y}$ represent different groups divided based on sensitive features and ground-truth labels.


\textbf{Iso-Index} is a metric used to assess the fairness and equity of information retrieval systems. It measures the isolation of certain groups within the retrieved results, indicating the extent to which specific groups are underrepresented or segregated in the search results. A lower Iso-Index value suggests less isolation and, therefore, a more equitable distribution of information across different groups. This metric is particularly important in contexts where diversity and fairness are critical, such as social media, job recommendations, and news aggregation.
\begin{equation*}
    \mathrm{ISO}\text{-}\mathrm{index} = \lambda \cdot \mathrm{Diversity} + (1-\lambda)\cdot \mathrm{Fairness},
\end{equation*}
where $\lambda$ is a hyper-parameter.

\textbf{Novelty} is a metric that favors recommendations of items that are different from those the user has already encountered. It emphasizes introducing fresh and less familiar content, rather than recommending similar or previously interacted items. Following~\cite{silveira2019good}, novelty is defined as the degree to which recommended items differ from the user’s past interactions, with a higher novelty value indicating more diverse and less repetitive recommendations:
\begin{equation*}
    \mathrm{Novelty}@k = \sum_{i=1}^{N} \sum_{j=1}^{|\widehat{L}^k_i|}\frac{\log(\mathrm{Pop}(\hat{l}_{i,j})+1)}{|\widehat{L}^k_i|},
\end{equation*}
where $\hat{l}_{i,j}$ is the $j$-th item in the ranking list $\widehat{L}^k_i$, $\mathrm{Pop}(\hat{l}_{i,j})$ denotes the popularity of item $\hat{l}_{i,j}$.


\subsubsection{Multi-Objective Optimization Metrics}
\label{sec:Multi-objective optimization metric}

When the task requirements involve multi-objective or multi-task scenarios, the optimization goals may conflict and have constraints among themselves. That is, the optimal solution is not a single point but a surface or curve. To measure the quality of the front generated by the control function, new evaluation metrics are required. However, due to the scarcity of work on controllable multi-objective optimization in the IR field, traditional multi-objective optimization evaluation metrics are rarely used in the IR domain. Here, we briefly introduce some evaluation metrics related to controllable multi-objective optimization.


According to~\cite{audet2021performance}, performance indicators in multiobjective optimization are mainly focus on  continuous Pareto front approximation. However, in the field of controllable 
multi-objective optimization, the Pareto front approximation is not obtained through evolutionary algorithms that yield a finite (discrete) set of points. For example,~\cite{cen2020controllable} achieves diversity control by adjusting the continuous value $\lambda$ in the aggregation module, and similarly,~\cite{chen2023controllable} controls diversity and accuracy by adjusting the continuous value $\lambda$ in the hypernetwork's input weights. Thus, we present the evaluation metrics for discrete Pareto front approximations:



\textbf{Hypervolume}\cite{guerreiro2020hypervolume} measures the volume of the objective space dominated by a set of solutions $S$, bounded by a reference point $z$. It evaluates both convergence and diversity. A larger HV value indicates better performance:
\begin{equation*}
\mathrm{HV} = \lambda \left( \bigcup_{i=1}^{|S|} \prod_{j=1}^m \left[f_j^{(i)}, z_j\right] \right),
\end{equation*}
where $S$ is the solution set, $f_j^{(i)}$ is the objective value of solution $i$ in the $j$-th objective, $z_j$ is the reference point in the $j$-th objective, and $\lambda(\cdot)$ denotes the Lebesgue measure.

\textbf{R2}~\cite{hansen1994evaluating} is a scalarization-based metric used to evaluate the quality of a solution set in multi-objective optimization. It measures the proximity of the solution set to an ideal set based on predefined weight vectors, without requiring the true Pareto front. This metric is defined as:
\begin{equation*}
    \mathrm{R2} = \frac{1}{|W|} \sum_{w \in W} \min_{x \in P} \sum_{i=1}^m w_i f_i(x),
\end{equation*}
where $ W $ is a set of weight vectors that represent user-defined preferences in the objective space, $ f_i(x) $ is the value of the $ i $-th objective function for a solution $ x $, $ P $ is the set of solutions being evaluated, and $ m $ is the number of objectives.

In some studies, the continuous Pareto front can be obtained, allowing the measurement of the distance between the discrete solution set and the Pareto front.

\textbf{Generational Distance}~\cite{van1998multiobjective} quantifies the convergence of the obtained solution set $P$ to the true Pareto front $P^*$. It computes the average Euclidean distance between each solution in $P$ and the nearest point in $P^*$: 
\begin{equation*}
\mathrm{GD} = \left( \frac{1}{|P|} \sum_{i=1}^{|P|} d_i^p \right)^{\frac{1}{p}},
\end{equation*}
where $|P|$ is the number of solutions in $P$, $d_i$ is the Euclidean distance from the $i$-th solution in $P$ to the nearest point in $P^*$, and $p$ is typically set to 2.

\textbf{Inverted Generational Distance}~\cite{coello2004study} evaluates both convergence and diversity by computing the average distance from each point in the true Pareto front $P^*$ to the nearest solution in the obtained set $P$: 
\begin{equation*}
\mathrm{IGD} = \frac{1}{|P^*|} \sum_{j=1}^{|P^*|} d_j,
\end{equation*} 
where $|P^*|$ is the number of points in $P^*$, and $d_j$ is the Euclidean distance from the $j$-th point in $P^*$ to the nearest solution in $P$.

\subsection{Datasets}
\label{sec:dataset} 
In this section, we summarize the commonly used datasets for controllable learning in information retrieval applications. To meet the various control requirements mentioned in Section~\ref{sec:what_to_control}, the datasets need to include the corresponding features.
For example, to control result diversity, the data needs to include category information of items; to protect user privacy or control user history, the data needs to include user profile and interaction history. Here, we summarize the commonly used publicly available datasets that can be used for controllable learning research:


\textbf{Amazon}~\cite{ni2019justifying,hou2024bridging}: 
This dataset comprises 142.8 million product reviews from various categories on Amazon\footnote{\url{https://www.amazon.com/}}, along with user and item profiles. And it includes category information of items, which can be used for multi-objective control such as diversity and fairness. It also contains time information, allowing for the extraction of users' historical sequences and subsequently controlling these historical sequences.

\textbf{Ali\_Display\_Ad\_Click}~\cite{zhou2018deep}\footnote{\url{https://tianchi.aliyun.com/dataset/56}}: The dataset includes records for 1 million users and 26 million ad display/click logs, featuring 8 user profile attributes (such as ID, age, and occupation) and 6 item features (such as ID, campaign, and brand).



\textbf{UserBehavior}~\cite{cen2020controllable}\footnote{\url{https://tianchi.aliyun.com/dataset/649}}: It collects user behaviors from Taobao’s recommender systems~\cite{zhu2018learning}. This dataset includes all behaviors (such as clicks, purchases, add-to-cart actions, and likes) of approximately one million randomly selected users with activity between November 25, 2017, and December 3, 2017.

\textbf{MovieLens}\footnote{\url{https://grouplens.org/datasets/movielens}}: This dataset is a classical movie recommendation dataset. 
It includes dataset versions of various sizes, such as 100k, 1M, 10M, and 20M. It includes information on users' gender, age, and occupation, as well as item category information.

\textbf{MS~MARCO}~\cite{nguyen2016ms}~(Microsoft Machine Reading Comprehension): 
This extensive dataset is designed for evaluating machine reading comprehension, retrieval, and question-answering capabilities in web search scenarios. It includes two benchmarks: document ranking and passage ranking, encompassing a total of 3.2 million documents and 8.8 million passages. Compiled from real user queries extracted from Microsoft Bing’s search logs, each query is paired with annotated relevant documents. This dataset spans a wide variety of question types and document genres, aiming to assess the performance of GR systems in complex web search scenarios.

In fact, there are currently no dedicated domain-specific datasets for controllable learning. Developing more tailored datasets remains an open challenge, which we leave for future work.


%% file: sections/6_challenge.tex
\section{Challenges in Controllable Learning}
In this section, we discussed the challenges that may arise when applying controllable learning methods in the domain of information retrieval.

\label{sec:challenges}




\subsection{Balancing Difficulty in Training}
\label{sec:balancing_difficulty}
The difficulty of balancing controllability with performance and efficiency is the pivotal challenge. Pursuing controllability often leads to a trade-off, potentially compromising performance or other user-centric optimization metrics and adversely impacting accuracy or user experience. For instance, controllability may be sought through the manipulation of hyperparameters—like a balancing factor in a loss function or a direct evaluative indicator. For instance, ComiRec~\cite{cen2020controllable} adjusts the balancing factor within its aggregation module to enhance diversity. Empirical evidence suggests that while diversity improves, accuracy can be compromised to some degree.
\subsection{Absence of Evaluation}
\label{sec:absence_of_evaluation}
The absence of standardized benchmarks and evaluation metrics also hinders the development of Controllable Learning for IR, likely attributed to the very beginning stage of controllability learning and the lack of consensus on such metrics. Despite the shared objective of amplifying recommendation diversity, methods like CCDF~\cite{zhang2024practical} and ComiRec~\cite{cen2020controllable} adopt disparate evaluative approaches—the former implictly measuring it by designing a specific scenario and leveraging Hit Ratio (HR), while the latter assesses the diversity of top N recommended items by analyzing inter-category difference. The assortment of perspectives on controllability and the consequent need for tailored evaluation metrics can prevent direct methodological comparisons and hinder the progression of the field.

\subsection{Setting Task Descriptions in Controllable Learning}
\label{sec:challenge:3}
In the context of a CL framework, the task target determines what is controllable, while the task description serves as the instructions given by humans to the learner. A crucial issue is how to set the task target and transform it into a human-understandable and precise description. Task descriptions are not limited to vectors or text; they can also take the form of images, graphs, rules, and other formats.

\subsection{Challenges in Online Environments}
\label{sec:online_challenges}
Scalability in real-world applications (e.g., IR systems), particularly those dealing with streaming data and requiring online learning, is a formidable challenge. While research on controllability has been extensive in offline environments, integrating these principles into streaming IR applications, like online learning and reinforcement learning (RL), is yet to be fully realized. For instance, Wang et al. has designed influence-function-based models tailored to user preferences~\cite{wang2024would}, but these models aren't equipped to handle the swift changes in preferences without undergoing a retraining process—impractical for the real-time demands of streaming applications. Consequently, there's an imperative need for research to pivot more resources into online settings. Future work could prioritize the development of models that can adjust on-the-fly to changing data and preferences. This involves creating systems capable of incremental learning and employing real-time feedback to refine their performance continuously. Progress in this direction will be a significant step toward the practical deployment of controllable learning models in dynamic streaming IR environments.


\section{Future Directions}
\label{sec:future_direction}
Controllable learning is gaining increasing attention, even though much of the current focus remains within implicit domains. This section highlights several future development directions, based on the challenging issues and emerging technologies (e.g., large language models). It is hoped that this chapter will provide valuable inspiration for future research.

\textbf{Why should we control? Theoretical analyses of controllable learning.}
Finding the optimal hypothesis in the hypothesis space for controllable learning is a more challenging task compared to traditional machine learning. Specifically, given the task requirements, the essence of in-processing controllable learning methods lies in establishing a mapping between the task target and the model parameters. Due to the vast parameter space of current deep learning models, uncovering structural information within this space and understanding causal associations with the target require rigorous theoretical analysis and effective training methods. This stands as a critical future research direction in controllable learning.

\textbf{Controllable sequential decision-making models.}
In numerous streaming applications such as streaming recommender systems, feedback often takes the form of bandit feedback (i.e., feedback on decisions not executed is unobserved). This presents new challenges for sequential decision-making models such as reinforcement learning and online learning. Balancing exploration and exploitation while achieving adaptive control over task requirements is a crucial issue in both the theoretical analysis and practical applications of controllable learning.

\textbf{Empowering LLM-based AIGC through controllable learning.} Existing controllable generation methods typically rely on large language models (LLMs), using natural language prompts to control the input and obtain controllable AI Generated Content (AIGC). However, exploring controllable learning techniques to manipulate model parameters or outputs for achieving more specific task targets (e.g., preferences across multiple objectives) remains an area requiring further investigation.

\textbf{Cost-effective control learning mechanisms.} 
As stated in the definition of controllable learning (Definition~\ref{def:CL:first}), 
compared to the original learner, a controllable learner requires additional assistance from control functions, which inevitably introduces additional computational cost. Given the substantial computational costs linked with large-scale models, investigating efficient and cost-effective control mechanisms becomes imperative for future research.

\textbf{Controllable learning for multi-task switching.}
Currently, in the field of information retrieval, there is limited research on controllable learning specifically tailored for search. Most existing work focuses on recommender systems. Therefore, adapting and extending controllable learning methods to the search represents a key direction for future research. One possible direction is how to utilize the same controllable matching model to adaptively switch between search and recommendation tasks.
More broadly, leveraging a small set of controllable learning models to address multi-task, multi-objective, and multi-scenario switching challenges will not only enhance flexibility in addressing varied task requirements but also drive the development of novel methodologies capable of adapting to dynamic environments.



\textbf{Demand for resource and metrics.} 
Despite its significance, controllable learning lacks dedicated datasets and standardized evaluation metrics. For instance, the collection or construction of labels or user feedback across multiple objectives or diverse task requirements is crucial for the training and testing of controllable learners. Addressing these gaps represents a pivotal area for future research.

%% file: sections/7_conclusion.tex
\section{Conclusion}
The landscape of controllable learning (CL) has been significantly enriched through the integration of diverse methodologies aimed at enhancing the trustworthiness of machine learning. CL's ability to dynamically adjust learner according to predefined targets and adapt without retraining when those targets evolve makes it a critical component in fostering reliable and adaptive machine learning models. Particularly within the context of information retrieval (IR), CL offers a means to address the complexity and dynamism inherent in information needs. 
However, there remains much to explore in terms of theoretical guarantees, computational efficiency, and the integration of LLMs into controllable learning. Further advancements in these areas could lead to more sophisticated, user-centric, and controllable AI models that meet the diverse and evolving needs of IR applications.

